\documentclass[lettersize,journal]{article}

\usepackage{amsmath,amsfonts}
\usepackage{algorithmic}
\usepackage{array}
\usepackage[caption=false,font=normalsize,labelfont=sf,textfont=sf]{subfig}
\usepackage{textcomp}
\usepackage{stfloats}
\usepackage{url}
\usepackage{verbatim}
\usepackage{graphicx}
\def\BibTeX{{\rm B\kern-.05em{\sc i\kern-.025em b}\kern-.08em
    T\kern-.1667em\lower.7ex\hbox{E}\kern-.125emX}}
\usepackage{balance}
\usepackage{xcolor}
\usepackage{algorithm}
\usepackage{algorithmic}

\usepackage{pgfplots} 
\usepackage{float} 
\usepackage{caption}

\begin{document}
\title{Asymmetric Momentum: A Rethinking of Gradient Descent}
\author{Gongyue Zhang, Dinghuang Zhang, Shuwen Zhao\\Donghan Liu, Carrie M. Toptan and Honghai Liu}


\maketitle

\begin{abstract}
Through theoretical and experimental validation, unlike all existing adaptive methods like Adam which penalize frequently-changing parameters and are only applicable to sparse gradients, we propose the simplest SGD enhanced method, Loss-Controlled Asymmetric Momentum(LCAM). By averaging the loss, we divide training process into different loss phases and using different momentum. It not only can accelerates slow-changing parameters for sparse gradients, similar to adaptive optimizers, but also can choose to accelerates frequently-changing parameters for non-sparse gradients, thus being adaptable to all types of datasets. We reinterpret the machine learning training process through the concepts of weight coupling and weight traction, and experimentally validate that weights have directional specificity, which are correlated with the specificity of the dataset. Thus interestingly, we observe that in non-sparse gradients, frequently-changing parameters should actually be accelerated, which is completely opposite to traditional adaptive perspectives. Compared to traditional SGD with momentum, this algorithm separates the weights without additional computational costs. It is noteworthy that this method relies on the network's ability to extract complex features. We primarily use Wide Residual Networks for our research, employing the classic datasets Cifar10 and Cifar100 to test the  ability for feature separation and conclude phenomena that are much more important than just accuracy rates. Finally, compared to classic SGD tuning methods, while using WRN on these two datasets and with nearly half the training epochs, we achieve equal or better test accuracy.
\\Our demonstration code is available at https://github.com/hakumaicc/Asymmetric-Momentum-LCAM
\end{abstract}


\section{Introduction}


In machine learning, optimizers implement gradient descent. Adaptive optimizers, represented by Adam\cite{kingma2014adam}, have been in competition with traditional optimizers SGD with momentum. From the initial AdaGrad\cite{duchi2011adaptive} to Rmsprop\cite{RmsProp}, and then to Adam and AmsGrad\cite{loshchilov2018decoupled}, adaptive optimizers have undergone many improvements. However, their applicability remains limited to sparse gradients. While adaptive optimizers can converge quickly on sparse datasets, their performance is still not as good as traditional SGD with momentum on non-sparse datasets. Similarly, although SGD can perform well under appropriate scheduling, its convergence is slower. This means that while adaptive optimizers may require twice the computational effort, SGD also needs twice the number of iterations. Consequently, both types of optimizers still play different roles on the datasets where they are most effective. This has also led to some discussions on switching between different optimizers, such as methods for switching from Adam to SGD\cite{keskar2017improving}.

The saddle point problem\cite{dauphin2014identifying} is the primary factor affecting the effectiveness of convergence in the early to mid-stages and also influences the final convergence location. This paper mainly discusses the saddle point issue. Mathematically, a saddle point can be described as a point in a multivariate function where it acts as a local maximum in certain directions and a local minimum in others. In high-dimensional spaces, due to the vast number of directions, finding a genuine local minimum becomes highly challenging, making saddle points more prevalent. The optimization surface of deep learning models is intricate, comprising many saddle points and flat regions.

This paper reinterpret the iterative process of machine learning optimizers and primarily discusses through experimentation. The contributions of this paper are as follows:

\begin{itemize}
	\item{We experimentally demonstrate that for non-sparse datasets, weights with larger changes at beginning should be accelerated, which is completely opposite to the perspective of traditional adaptive optimizers. This implies that improving and enhancing adaptive optimizers solely based on historical gradient information is incompleteness.}
	\item{We make breakthrough that use loss values for weight separation and design saddle-point experiments to compare the effects brought by different directions of weight acceleration, validating that different datasets should accelerate weights in different directions.}
\end{itemize}

%

It should be noted that this paper focuses more on exploring the fundamental theory of machine learning but not final accuracy. Our experiments are based on the Wide Residual Network\cite{zagoruyko2016wide}. In our preliminary tests, other networks showed similar effects in the special experimental part, which can also prove the correctness of the theory. However, it may not be effectively applied to the actual training of all networks. This is because the Wide Residual Network has a strong ability to distinguish between sparse and non-sparse features. Finally, we simply employ a very straightforward scheduling approach to validate the effectiveness of Asymmetric Momentum in Wide Residual Network.

\section{Related Work}

\subsection{SGD}

Stochastic Gradient Descent (SGD)\cite{robbins1951stochastic} is a variant of the gradient descent optimization method for minimizing an objective function that is written as a sum of differentiable functions. Instead of performing computations on the whole dataset, which can be computationally expensive for large datasets, SGD selects a random subset (or a single data point) to compute the gradient of the function. Momentum is introduced to the vanilla SGD to make the updates more stable and to dampen oscillations. It takes into account the past gradients to smooth out the updates.
\subsection{Adaptive Optimizer}
Adaptive optimizers,such as AdaGrad, RMSprop, Adam, AmsGrad and AdamW\cite{loshchilov2018decoupled}, typically accumulate past gradients, with different adaptive optimizers employing varied strategies for accumulating historical gradients. Generally, they use accumulated gradients to penalize parameter changes, thereby dynamically adjusting the learning rate for each weight and facilitating faster model convergence.

\section{Loss-Controlled Asymmetric Momentum(LCAM)}

In this section, we first introduce the concepts of weight coupling and weight traction to provide a simplified explanation of the training process of machine learning. We discuss the general algorithms of adaptive optimizers and speculate on the limitations of adaptive optimizers, specifically their ability to perform well only in the context of sparse gradients. We propose an improved SGD method, Loss-Controlled Asymmetric Momentum(LCAM), which allows for arbitrary combinations of momentum to accelerate in sparse or non-sparse directions. We implement control methods through the oscillatory properties of the loss value, and provide details of loss behavior during the training process.


\subsection{Weight Coupling}

Firstly, we introduce the concept of coupling. Representing the input set as $X$ and the output label as $Y$, although the weight segment is quite complex, we believe that it can ultimately be represented by multiple weight matrices ($\Theta$) multiplying together. We categorize these weights $\Theta$ into three types, $\Theta_c$, $\Theta_s$ and $\Theta_n$. Among them:

%

$\Theta_c$ represents constant weights that are less likely to be influenced by other parameters.

$\Theta_n$ signifies \textbf{non-sparse} weights that are easily trained, reflecting more prominent features from the raw data and undergo significant changes during training.

$\Theta_s$ stands for \textbf{sparse} weights that are hard to train,  capturing features that are harder to extract from the raw data, changing minimally throughout the training process. 
Thus, we can break it down and represent it as: 

\begin{align}
	Y = (\prod_{j}^{c}\Theta_{j\in c}\prod_{k}^{n}\Theta_{k\in n}\prod_{l}^{s}\Theta_{l\in s}+b)\times X
\end{align}

\subsection{Weight Traction}
Let's now re-evaluate our understanding of the machine learning training process. In the initial stages of training, since the weights are severely offset, changes in individual weights won't cause significant oscillations in their interrelations. Once the model enters an oscillatory state, given that $\Theta_c$, $X$ and $Y$ essentially remains in a quasi-fixed state, it will only produce a certain level of noise from the stochastic gradient output. Thus, $\Theta_n$ and $\Theta_s$ would essentially be in a coupling state, causing both weights groups to oscillate around the optimal point $Minima$, with a certain fixed value as their center.

\begin{figure}[H]
	\centering
	\includegraphics[width=2.5in]{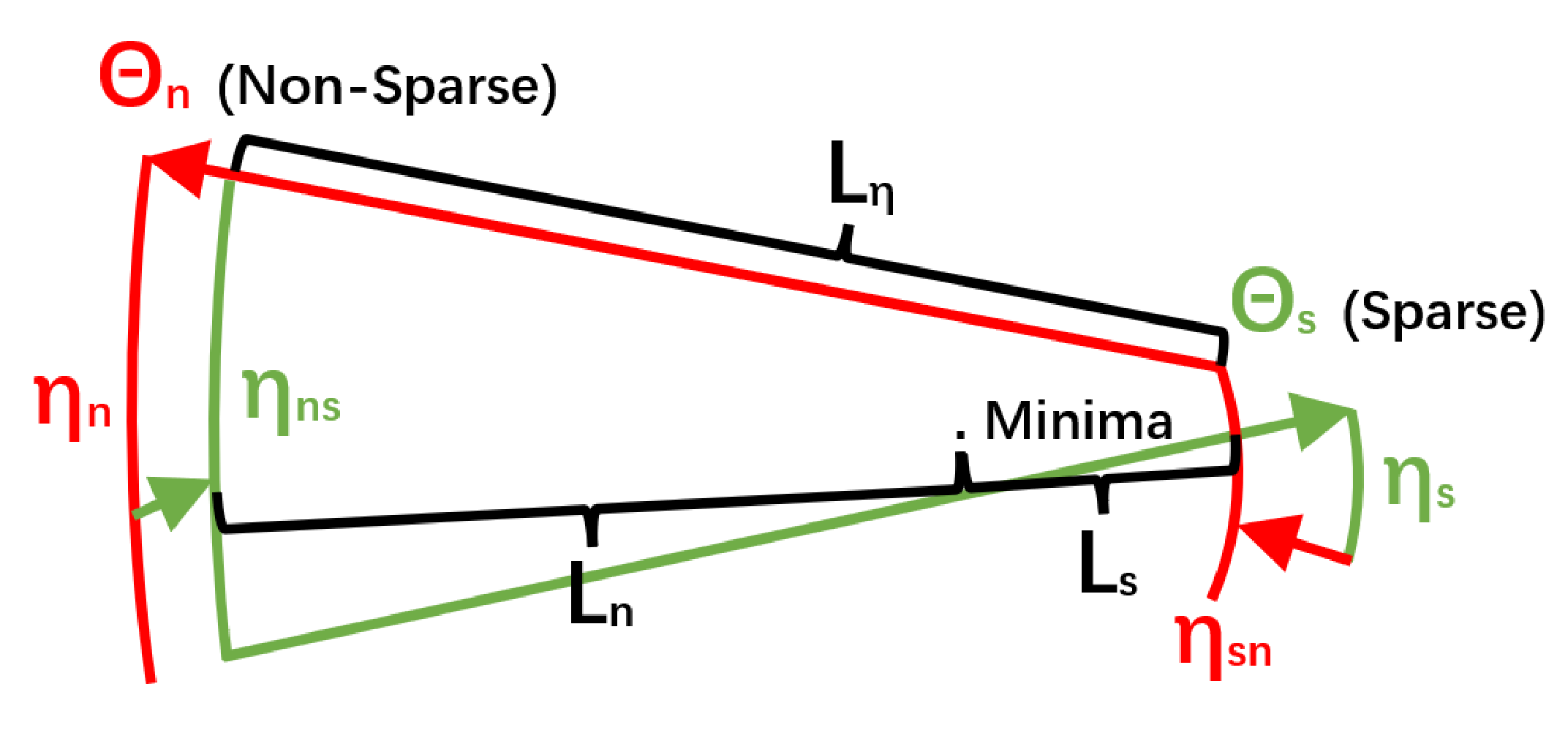}
	\caption{A simple demonstration of the weight coupling state of two groups parameters, $\Theta_n$ and $\Theta_s$.}
	\label{png0301}
\end{figure}

We believe that the \textbf{Weight Coupling} results in what we term as \textbf{Weight Traction Effect}, where the oscillation amplitude is primarily influenced by the learning rate. As illustrated in the figure.\ref{png0301}, we demonstrate the phenomena arising from the mutual traction between $\Theta_n$ and $\Theta_s$. The red represents changes dominated by $\Theta_n$, while the green indicates changes dominated by $\Theta_s$. Let's denote the traction force between the weights as $L_\eta$. We can conceptualize $L_\eta$ as a 'traction rod'. This force is influenced by the learning rate. Evidently, the larger the learning rate, the stronger the traction force. The total \textbf{Loss} can be described as:

\begin{align}
	Loss_{Total}=Loss_{(L_c,L_\eta)}&, 	L_\eta=Loss_{(L_n,L_s)}
\end{align}

Therefore, the training process can be understood as being comprised of two distinct parts.

\begin{itemize}
	\item \textbf{Non-Sparse Quick-Changing Weight Group $\Theta_n$:} The red part is dominated by $\Theta_n$, where $\Theta_n$ advances by $\eta_n$ through the learning rate. During this phase, $\Theta_n$ exerts a pull on $\Theta_s$ to the left, changing $\Theta_s$'s advance from $\eta_s$ to $\eta_{sn}$.
	\item \textbf{Sparse Slow-Changing Weight Group $\Theta_s$:} Similarly, the green part is led by $\Theta_s$, wherein $\Theta_s$ moves forward by $\eta_s$ through the learning rate. In this phase, $\Theta_s$ pulls $\Theta_n$ to the right, causing $\Theta_n$'s progression to shift from $\eta_n$ to $\eta_{ns}$
\end{itemize}

\subsection{Limitation of Adaptive Optimizers}

In reality, the inner workings of machine learning are very complex. For the sake of simplicity and ease of explanation, we have simplified the training process through the above methods. Next, let's consider the algorithms of adaptive optimizers, and similarly simplify them for ease of discussion. We will start by introducing AdaGrad, followed by a discussion on Adam and fixed-momentum SGD.

For AdaGrad\cite{duchi2011adaptive}, the parameter update rule is:

\begin{align}
	&g_t=\nabla J(\Theta_t)\\
	&G_t=G_{t-1}+g_t \odot g_t\\
	&\Theta_{t+1}=\Theta_t-\frac{\eta}{\sqrt{G_t+\epsilon}} \odot g_t
\end{align} 

$g_t$ is the gradient at time step $t$, $\nabla J$ is the gradient of the objective function $J$ at $\Theta_t$, $G_t$ is a diagonal matrix where each diagonal entry $G_t^{ii}$ is the sum of the squares of the gradients with respect to $\Theta^{ii}$ up to time step $t$. The weights $\Theta$ are updated by subtracting the learning rate $\eta$ divided by the square root of $G_t+\epsilon$ element-wise multiplied by the gradient $g_t$.

As we can see, the core idea behind adaptive optimization is to accumulate historical gradients $G_t$. In the initial stages, it allows for rapid updates of the weights while keeping track of the gradients. In the later stages, the accumulated historical gradients $\sqrt{G_t+\epsilon}$ are used to slow down the update of weights that were updated quickly earlier on, while continuing to train the side of the weights that have been updating more slowly.

\begin{figure}[H]
	\centering
	\includegraphics[width=5in]{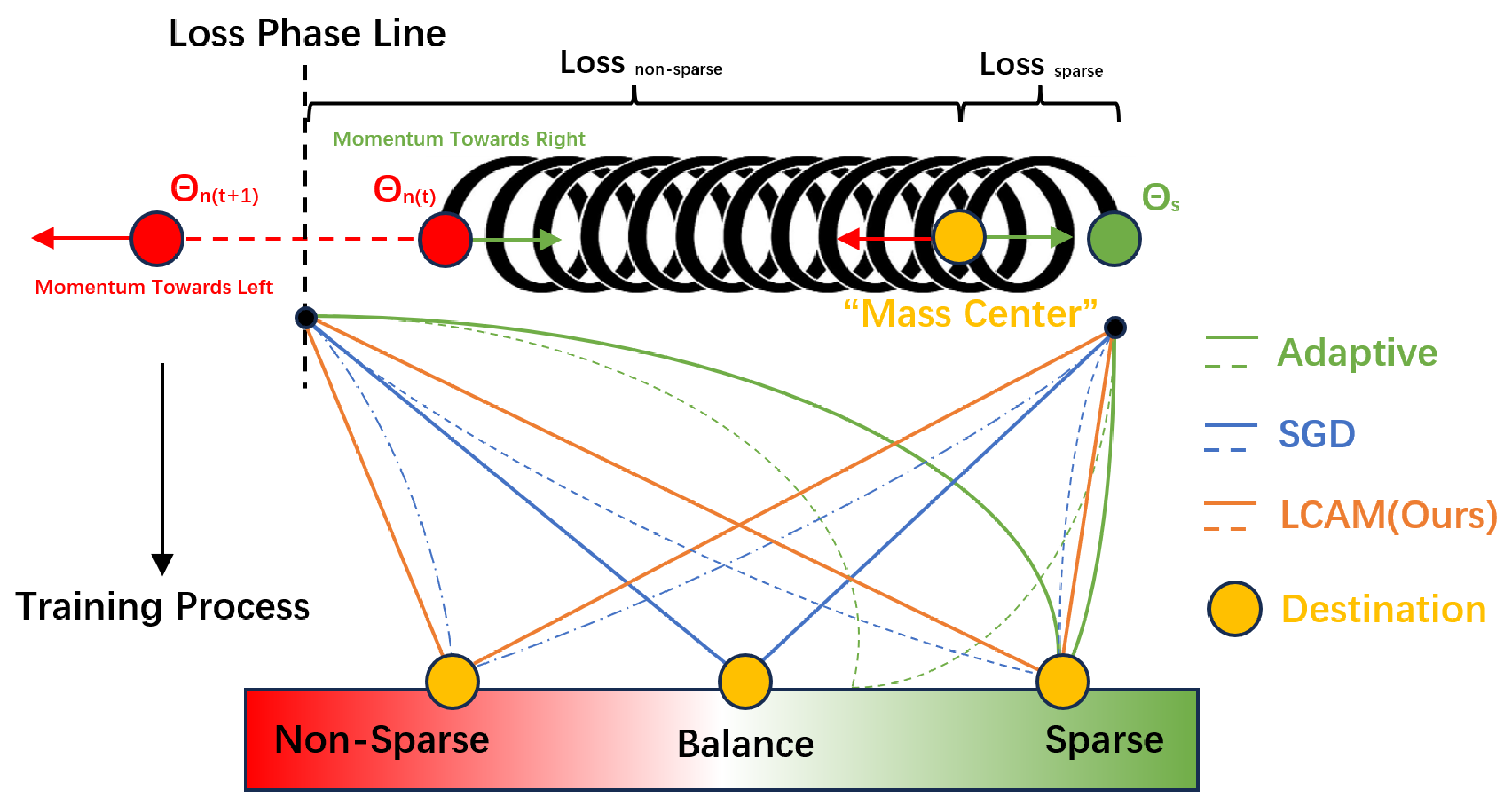}
	\caption{$\Theta_n$ and $\Theta_s$ serve as parameters on both sides, undergoing oscillations similar to that of a spring. The adaptive optimizer AdaGrad essentially accumulates historical gradients to quickly reduce $\Theta_n$, thereby weakening the spring's effect and prematurely reducing the non-sparse parameters in the direction of $\Theta_n$. This implies that it will be pushed towards the sparse direction, as indicated by the green line. Even the improved represented by the green dashed line, Adam, cannot fully correct the early-stage bias. As shown by the blue line, while SGD can be effective on balanced datasets, it requires a long time to change direction and escape saddle points for both sparse and non-sparse datasets. Our method, LCAM, represented by the orange line, leverages the characteristics of loss oscillation to determine the acceleration ratios of parameters in different directions from the beginning, thus adopting corresponding directional strategies under various datasets.}
	\label{png0302}
\end{figure}

It's now time to introduce another weight-traction model. To make it more illustrative, we will use the behavior of a \textbf{Spring} to analogize the process of machine learning. Just like in Figure.\ref{png0301}, we introduce Figure .\ref{png0302} also represents the traction effect between $\Theta_n$ and $\Theta_s$. Consistent with the core idea of adaptive optimization, $\Theta_n$ represents parameters that have historically changed quickly in the non-sparse direction; $\Theta_s$ represents parameters that have historically changed slowly in the sparse direction. The length of the spring represents the learning rate $\eta$. When the learning rate $\eta$ is large, oscillations will be more pronounced; whereas when the spring length is very small, the amplitude of oscillations will also decrease.

In the context of a spring system, we can think of $\Theta_s$ as having a larger mass, because it changes very little when influenced by the learning rate, or in this case, external forces. Conversely, $\Theta_n$ has a smaller mass, resulting in large oscillations when subjected to external forces. In terms of the adaptive optimization concept, what is primarily tracked is the variation in $\Theta_n$, and the length of the spring is rapidly reduced accordingly. This effectively isolates $\Theta_n$, allowing it to quickly approach $\Theta_s$ and subsequently update together in the later stages.

\textbf{Why do adaptive optimizers perform well only on sparse datasets?} Now, we can explain this phenomenon. By penalizing $\Theta_n$, which represents the weights in the non-sparse gradient direction, it inevitably cause the mass center of the spring system, or the overall weights, to shift significantly towards the sparse direction in the bottom right as the gradients update. This is represented by the process denoted by the green solid line in Figure.\ref{png0302}.

This is also why Adam's performance tends to be better than that of AdaGrad. If we refer to Adam's\cite{kingma2014adam} update mechanism:

\begin{align}
	&g_t=\nabla J(\Theta_t)\\
	&m_t=\beta_1 \times m_{t-1}+(1-\beta_1)\times g_t\\
	&v_t=\beta_2 \times v_{t-1}+(1-\beta_2)\times g_t^2\\
	&\hat{m_t}=\frac{m_t}{1-\beta_1^t}\\
	&\hat{v_t}=\frac{v_t}{1-\beta_2^t}\\
	&\Theta_{t+1}=\Theta_t-\frac{\eta}{\sqrt{\hat{v_t}+\epsilon}}\times \hat{m_t}
\end{align} 

Clearly, Adam allows the decay $\beta_2$ for early recorded gradients, which gives the direction of $\Theta_n$ the opportunity to adjust in the later stages and potentially shift its destination to the left or right. However, from our perspective, Adam doesn't solve all the issues, as its weight updates are still aggressive in the early stages. This means that later adjustments cannot fully correct the aggressive behavior of shrinking the learning rate in its early phases. As a result, its final destination still leans towards the bottom right, as indicated by the green dashed line in Figure.\ref{png0302}. Although its performance is significantly better than AdaGrad, its versatility is still lacking, and not sufficient to compare with a balanced SGD in non-sparse gradient, which is inherently closer to the optimal point for datasets like Cifar10.

In other words, even fixed-momentum SGD is not fast enough to reach the optimal point for a non-sparse dataset such as Cifar10. Because SGD itself is just a balanced approach, it doesn't have a distinct directional bias, as indicated by the blue line in Figure.\ref{png0302}. This means that SGD still needs to make turns during training based on the final destination direction, which results in a waste of time. For SGD, whether it's $\Theta_n$ or $\Theta_s$, depending on the dataset, one of them will inevitably reach the destination ahead of the other. This makes SGD more susceptible to getting stuck at saddle points and must rely on long training duration to allow $\Theta_n$ and $\Theta_s$ to pull each other out of the saddle points, thereby expanding its range of reachable destination.

This is also why Adam appears more effective, its decay exponential moving average strategy allows it to cover a broader range of destination, training without the need for extra training time, even though this coverage is entirely bias towards the sparse direction.

Under our theoretical framework, all traditional gradient descent optimizers are incomplete. Theoretically speaking, there is no optimizer that can cover the full range of gradient types.

\subsection{Asymmetric Momentum}

To address the issue of universally destination range applicable, we propose Loss-Controlled Asymmetric Momentum(LCAM). By averaging the loss, we divide training process into different loss phases and using different momentum. We can isolate weights without additional computational cost compared to traditional SGD with momentum. This not only allows for accelerating weights that change slowly in the sparse gradients but also weights that change frequently in non-sparse gradients, thereby making it adaptable to all kinds of datasets. It's important to note that this method relies on the network's ability to extract features of varying complexity.

Considering $\Theta_n$ in Figure.\ref{png0302}, in equations (2), we reveal that the loss is composed of three parts, among which $L_c$ is constant. Taking into account the spring system, $\Theta_s$ doesn't change much, so the corresponding $L_s$ also changes very little. Therefore, the apparent changing $Loss_{Total}$ is mostly provided by $L_n$ with a direct ratio:

\begin{align}
	&\Delta Loss_{Total} \approx k \times \Delta L_n
\end{align} 

This means that for the lighter side of the spring, $\Theta_n$, its position can be easily determined by comparing the value of the loss with the average loss. Specifically, when the loss is larger, it will be on the left side of the loss phase line; when the loss is smaller, it will be on the right side of the loss phase line, showed in Figure.\ref{png0302}.

Now consider the effect of increasing the momentum at different phases. If we increase the momentum whenever $\Theta_n$ moves to the left side, it's clear that, with the accumulation of applied force, we will ultimately pull the mass center of the spring system towards the left, which is the red direction of non-sparse gradients. Conversely, if we increase the momentum whenever $\Theta_n$ moves to the right side, then as the force accumulates, we will eventually push the mass center of the spring system towards the right, which corresponds to the green direction of sparse gradients.

\begin{algorithm} 
	\caption{Loss-Controlled Asymmetric Momentum} 
	\label{alg1} 
	\begin{algorithmic}
		\REQUIRE Learning Rate $\eta$, Momentum $\beta$
		\REQUIRE Loss $L(t)$, Current Mean Loss $L_m(t)$
		\STATE $\eta \gets 0.1$ (Initialize Learning Rate)
		\WHILE{Iteration}
		\IF{$Epoch \textgreater 30$}
		\STATE $\eta \gets \eta \times 0.99985$
		\ENDIF
		\STATE Get Loss $L(t)$ in Current Iteration
		\STATE Optimizer $SGD(\eta, \beta)$ Update
		\STATE Calculate Current Mean Loss $L_m(t)$
		\IF{$L(t) \textless L_m(t)$}
		\STATE $\beta \gets 0.9$ (Sparse Momentum at Right Side)
		\ELSE
		\STATE $\beta \gets 0.95$ (Non-Sparse Momentum at Left Side)
		\ENDIF
		\ENDWHILE
	\end{algorithmic} 
\end{algorithm}

The method is very straightforward, and the algorithm is as shown in the Algorithm.\ref{alg1}. Note that this algorithm is applied in the final training algorithm for Cifar10, not the algorithm brought about by special experiments. For details on special experiments, please refer to the experiment section. During each iteration, the algorithm compares the current loss with the average loss to determine the current position of $\Theta_n$, thereby deciding whether to apply additional momentum.

\subsection{Limitation}

Returning to the initial problem of saddle points, the issue is essentially that $\Theta_n$ and $\Theta_s$ do not reach the optimal destination together. Traditional methods either extend the training time to allow $\Theta_n$ and $\Theta_s$ to pull each other out of the saddle points, or use adaptive method like Adam to leap over them. Do we eliminate the saddle point problem, directly by using asymmetric momentum to bring $\Theta_n$ and $\Theta_s$ to the optimal destination together?

No, this is only an optimistic question. In reality, we are only weaken the saddle point phenomenon. Machine learning itself is a highly complex process with high dimensions and not simply the movement of two variables. All the discussions above are based on simplified model, meaning that both $\Theta_n$ and $\Theta_s$ have groups of individual $\Theta_n(i)$ or $\Theta_s(j)$ each requiring different momentum, and we can only approximate their general trends. Furthermore, for networks with poorer feature extraction capabilities, asymmetric momentum may not effectively separate weights due to the mixture of weights, resulting in insufficient apparent oscillations brought about by $\Theta_n$ and $\Theta_s$. The key is we pointed out the improving direction of optimizers in future.

\section{Experiment}

We designed a very simple experiment to demonstrate the direction specificity of gradients. We tested using WRN28-10\cite{zagoruyko2016wide}, which is from original Residual Network\cite{he2016deep}, with the classic Cifar10 and Cifar100\cite{krizhevsky2009learning} as test datasets. Both datasets have 50,000 training samples, but one has ten categories while the other has a hundred. This implies that, in terms of data sample structure, the sparsity level of Cifar100 is ten times that of Cifar10. All experiments are based on SGD, and no dropout was used during testing. Hyperparameters were set with an initial learning rate of 0.1, momentum of 0.9, and weight decay set to $5\times 10^{-4}$, with a batch size of 128. To more easily test the impact of varying momentum on training, we shortened the drop nodes to epochs 30, 60, and 90, reducing the learning rate to 20\% of its value at these epochs. The rapid decline in learning rate deliberately traps the weights in a saddle point, serving as a baseline for discussing asymmetric momentum.

%

\subsection{Cifar10}

First, we conducted experiments on Cifar10. We used a total of four groups of momentum:

\begin{itemize}
	
	\item The first and second groups maintain a momentum of 0.9 and 0.95 throughout, serving as the baseline. These are represented by black and blue lines, respectively.
	
	\item The third group uses a momentum of 0.95 during the sparse phase on the right and 0.9 during the non-sparse phase on the left, to accelerate the sparse $\Theta_s$ side and push the weights to the right. This is represented by a green line.
	
	\item The fourth group uses a momentum of 0.95 during the non-sparse phase on the left and 0.9 during the sparse phase on the right, to accelerate the non-sparse $\Theta_n$ side and push the weights to the left. This is represented by a red line.
	
\end{itemize}

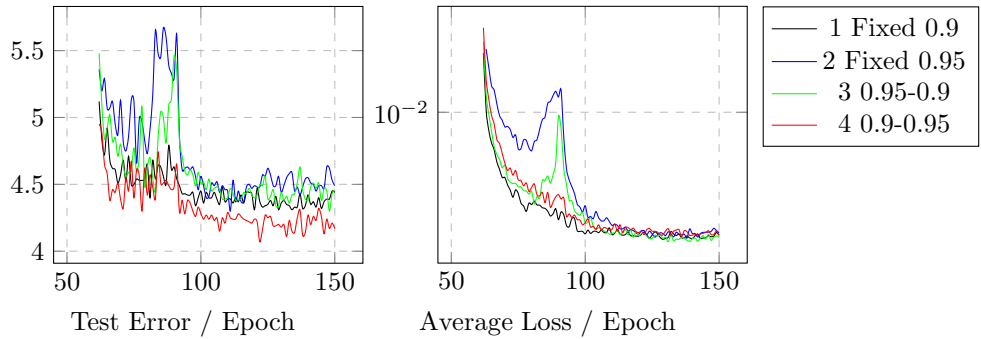
\begin{figure*}[ht]
	\centering
	\begin{minipage}[ht]{0.4\textwidth}
		\centering
		\begin{tikzpicture} 
			\begin{axis}[
				scale=0.6,
				tick align=inside, 
				xtick = {50,100,150},
				xmin = 45,
				grid style={dashed},
				xmajorgrids =true,
				ymajorgrids =true,
				legend style={at={(1,1)},anchor=north} 
				]
				
					%
				\addplot[smooth,black] plot coordinates { 
					(62, 5.119999885559082)(63, 4.840000152587891)(64, 4.71999979019165)(65, 4.920000076293945)(66, 4.679999828338623)(67, 4.619999885559082)(68, 4.610000133514404)(69, 4.510000228881836)(70, 4.570000171661377)(71, 4.679999828338623)(72, 4.519999980926514)(73, 4.710000038146973)(74, 4.5)(75, 4.5)(76, 4.539999961853027)(77, 4.53000020980835)(78, 4.53000020980835)(79, 4.539999961853027)(80, 4.650000095367432)(81, 4.400000095367432)(82, 4.679999828338623)(83, 4.539999961853027)(84, 4.639999866485596)(85, 4.630000114440918)(86, 4.559999942779541)(87, 4.5)(88, 4.789999961853027)(89, 4.610000133514404)(90, 4.630000114440918)(91, 4.539999961853027)(92, 4.489999771118164)(93, 4.429999828338623)(94, 4.429999828338623)(95, 4.420000076293945)(96, 4.420000076293945)(97, 4.440000057220459)(98, 4.340000152587891)(99, 4.400000095367432)(100, 4.449999809265137)(101, 4.360000133514404)(102, 4.440000057220459)(103, 4.400000095367432)(104, 4.400000095367432)(105, 4.440000057220459)(106, 4.389999866485596)(107, 4.349999904632568)(108, 4.480000019073486)(109, 4.369999885559082)(110, 4.389999866485596)(111, 4.389999866485596)(112, 4.360000133514404)(113, 4.400000095367432)(114, 4.369999885559082)(115, 4.420000076293945)(116, 4.360000133514404)(117, 4.440000057220459)(118, 4.440000057220459)(119, 4.429999828338623)(120, 4.309999942779541)(121, 4.46999979019165)(122, 4.369999885559082)(123, 4.329999923706055)(124, 4.340000152587891)(125, 4.349999904632568)(126, 4.380000114440918)(127, 4.340000152587891)(128, 4.409999847412109)(129, 4.309999942779541)(130, 4.360000133514404)(131, 4.380000114440918)(132, 4.349999904632568)(133, 4.340000152587891)(134, 4.309999942779541)(135, 4.329999923706055)(136, 4.329999923706055)(137, 4.429999828338623)(138, 4.320000171661377)(139, 4.340000152587891)(140, 4.420000076293945)(141, 4.349999904632568)(142, 4.369999885559082)(143, 4.329999923706055)(144, 4.400000095367432)(145, 4.320000171661377)(146, 4.369999885559082)(147, 4.389999866485596)(148, 4.389999866485596)(149, 4.449999809265137)(150, 4.440000057220459)
					
				};
				
				\addplot[smooth,blue] plot coordinates { 
					(62, 5.360000133514404)(63, 5.199999809265137)(64, 5.289999961853027)(65, 5.070000171661377)(66, 5.070000171661377)(67, 5.119999885559082)(68, 4.949999809265137)(69, 4.869999885559082)(70, 5.130000114440918)(71, 4.820000171661377)(72, 4.75)(73, 4.860000133514404)(74, 5.130000114440918)(75, 5.119999885559082)(76, 4.659999847412109)(77, 4.989999771118164)(78, 4.96999979019165)(79, 4.829999923706055)(80, 4.730000019073486)(81, 4.909999847412109)(82, 5.159999847412109)(83, 5.650000095367432)(84, 5.480000019073486)(85, 5.449999809265137)(86, 5.670000076293945)(87, 5.599999904632568)(88, 5.389999866485596)(89, 5.329999923706055)(90, 5.289999961853027)(91, 5.619999885559082)(92, 4.860000133514404)(93, 4.610000133514404)(94, 4.619999885559082)(95, 4.630000114440918)(96, 4.610000133514404)(97, 4.630000114440918)(98, 4.559999942779541)(99, 4.460000038146973)(100, 4.559999942779541)(101, 4.489999771118164)(102, 4.400000095367432)(103, 4.409999847412109)(104, 4.510000228881836)(105, 4.489999771118164)(106, 4.519999980926514)(107, 4.440000057220459)(108, 4.489999771118164)(109, 4.460000038146973)(110, 4.449999809265137)(111, 4.300000190734863)(112, 4.53000020980835)(113, 4.369999885559082)(114, 4.449999809265137)(115, 4.369999885559082)(116, 4.369999885559082)(117, 4.480000019073486)(118, 4.5)(119, 4.46999979019165)(120, 4.489999771118164)(121, 4.480000019073486)(122, 4.480000019073486)(123, 4.570000171661377)(124, 4.559999942779541)(125, 4.619999885559082)(126, 4.570000171661377)(127, 4.489999771118164)(128, 4.489999771118164)(129, 4.559999942779541)(130, 4.440000057220459)(131, 4.489999771118164)(132, 4.550000190734863)(133, 4.579999923706055)(134, 4.590000152587891)(135, 4.449999809265137)(136, 4.480000019073486)(137, 4.420000076293945)(138, 4.480000019073486)(139, 4.539999961853027)(140, 4.550000190734863)(141, 4.510000228881836)(142, 4.5)(143, 4.449999809265137)(144, 4.489999771118164)(145, 4.489999771118164)(146, 4.46999979019165)(147, 4.630000114440918)(148, 4.619999885559082)(149, 4.53000020980835)(150, 4.489999771118164)

				};
				
				\addplot[smooth,green] plot coordinates { 
					(62, 5.480000019073486)(63, 5.099999904632568)(64, 4.840000152587891)(65, 4.920000076293945)(66, 5.019999980926514)(67, 4.820000171661377)(68, 4.849999904632568)(69, 4.710000038146973)(70, 4.800000190734863)(71, 4.639999866485596)(72, 4.559999942779541)(73, 4.590000152587891)(74, 4.599999904632568)(75, 4.769999980926514)(76, 4.590000152587891)(77, 4.710000038146973)(78, 5.079999923706055)(79, 4.489999771118164)(80, 4.630000114440918)(81, 4.699999809265137)(82, 4.550000190734863)(83, 4.449999809265137)(84, 4.840000152587891)(85, 5.039999961853027)(86, 5.0)(87, 4.809999942779541)(88, 5.090000152587891)(89, 5.139999866485596)(90, 5.460000038146973)(91, 5.340000152587891)(92, 4.829999923706055)(93, 4.710000038146973)(94, 4.599999904632568)(95, 4.570000171661377)(96, 4.610000133514404)(97, 4.570000171661377)(98, 4.489999771118164)(99, 4.590000152587891)(100, 4.510000228881836)(101, 4.519999980926514)(102, 4.460000038146973)(103, 4.539999961853027)(104, 4.449999809265137)(105, 4.46999979019165)(106, 4.480000019073486)(107, 4.460000038146973)(108, 4.389999866485596)(109, 4.389999866485596)(110, 4.400000095367432)(111, 4.46999979019165)(112, 4.429999828338623)(113, 4.380000114440918)(114, 4.389999866485596)(115, 4.389999866485596)(116, 4.440000057220459)(117, 4.429999828338623)(118, 4.46999979019165)(119, 4.420000076293945)(120, 4.400000095367432)(121, 4.420000076293945)(122, 4.5)(123, 4.489999771118164)(124, 4.559999942779541)(125, 4.389999866485596)(126, 4.409999847412109)(127, 4.460000038146973)(128, 4.340000152587891)(129, 4.340000152587891)(130, 4.440000057220459)(131, 4.380000114440918)(132, 4.409999847412109)(133, 4.360000133514404)(134, 4.420000076293945)(135, 4.429999828338623)(136, 4.510000228881836)(137, 4.630000114440918)(138, 4.460000038146973)(139, 4.389999866485596)(140, 4.380000114440918)(141, 4.480000019073486)(142, 4.510000228881836)(143, 4.46999979019165)(144, 4.510000228881836)(145, 4.380000114440918)(146, 4.46999979019165)(147, 4.449999809265137)(148, 4.300000190734863)(149, 4.429999828338623)(150, 4.460000038146973)
					
				};
				
				\addplot[smooth,red] plot coordinates { 
					(62, 4.949999809265137)(63, 4.889999866485596)(64, 4.650000095367432)(65, 4.579999923706055)(66, 4.380000114440918)(67, 4.460000038146973)(68, 4.440000057220459)(69, 4.510000228881836)(70, 4.389999866485596)(71, 4.320000171661377)(72, 4.550000190734863)(73, 4.449999809265137)(74, 4.670000076293945)(75, 4.400000095367432)(76, 4.610000133514404)(77, 4.559999942779541)(78, 4.300000190734863)(79, 4.619999885559082)(80, 4.420000076293945)(81, 4.570000171661377)(82, 4.480000019073486)(83, 4.389999866485596)(84, 4.739999771118164)(85, 4.510000228881836)(86, 4.510000228881836)(87, 4.53000020980835)(88, 4.489999771118164)(89, 4.460000038146973)(90, 4.480000019073486)(91, 4.650000095367432)(92, 4.289999961853027)(93, 4.380000114440918)(94, 4.28000020980835)(95, 4.380000114440918)(96, 4.360000133514404)(97, 4.300000190734863)(98, 4.260000228881836)(99, 4.329999923706055)(100, 4.28000020980835)(101, 4.239999771118164)(102, 4.269999980926514)(103, 4.239999771118164)(104, 4.269999980926514)(105, 4.230000019073486)(106, 4.25)(107, 4.150000095367432)(108, 4.25)(109, 4.210000038146973)(110, 4.289999961853027)(111, 4.25)(112, 4.239999771118164)(113, 4.239999771118164)(114, 4.28000020980835)(115, 4.260000228881836)(116, 4.239999771118164)(117, 4.230000019073486)(118, 4.21999979019165)(119, 4.210000038146973)(120, 4.239999771118164)(121, 4.21999979019165)(122, 4.070000171661377)(123, 4.159999847412109)(124, 4.269999980926514)(125, 4.25)(126, 4.260000228881836)(127, 4.190000057220459)(128, 4.190000057220459)(129, 4.199999809265137)(130, 4.179999828338623)(131, 4.199999809265137)(132, 4.110000133514404)(133, 4.230000019073486)(134, 4.150000095367432)(135, 4.139999866485596)(136, 4.25)(137, 4.309999942779541)(138, 4.119999885559082)(139, 4.139999866485596)(140, 4.269999980926514)(141, 4.199999809265137)(142, 4.260000228881836)(143, 4.230000019073486)(144, 4.320000171661377)(145, 4.260000228881836)(146, 4.130000114440918)(147, 4.239999771118164)(148, 4.170000076293945)(149, 4.210000038146973)(150, 4.170000076293945)
					
				};
				
			\end{axis}	
		\end{tikzpicture}
		\\Test Error / Epoch
	\end{minipage}\begin{minipage}[ht]{0.4\textwidth}
		\centering
		\begin{tikzpicture} 
			\begin{axis}[
				scale=0.6,
				tick align=inside, 
				xtick = {50,100,150},
				xmin = 45,
				ymode=log,
				grid style={dashed},
				xmajorgrids =true,
				ymajorgrids =true,
				legend style={at={(1.4,1)},anchor=north} 
				]
				
					%
				
				\addplot[smooth,black] plot coordinates { 
					(62, 0.02453550510108471)(63, 0.010399635881185532)(64, 0.007588568609207869)(65, 0.006313718855381012)(66, 0.005677275359630585)(67, 0.0047197407111525536)(68, 0.004230706952512264)(69, 0.0037336968816816807)(70, 0.003731867764145136)(71, 0.0033351194579154253)(72, 0.003088788129389286)(73, 0.002768024103716016)(74, 0.003095302963629365)(75, 0.0029315631836652756)(76, 0.0025345764588564634)(77, 0.002425817074254155)(78, 0.0022339848801493645)(79, 0.002533394144847989)(80, 0.0024787564761936665)(81, 0.0026094361674040556)(82, 0.00246482458896935)(83, 0.002457213122397661)(84, 0.002363181672990322)(85, 0.0023388357367366552)(86, 0.0022425365168601274)(87, 0.0020861404482275248)(88, 0.0021530415397137403)(89, 0.0023298701271414757)(90, 0.0019723516888916492)(91, 0.002219761023297906)(92, 0.002119454788044095)(93, 0.001807881984859705)(94, 0.001769778085872531)(95, 0.002001335844397545)(96, 0.00200852588750422)(97, 0.001671929145231843)(98, 0.0016514973249286413)(99, 0.001653834362514317)(100, 0.0015907962806522846)(101, 0.0016840060707181692)(102, 0.0016019833274185658)(103, 0.001649810466915369)(104, 0.0016571453306823969)(105, 0.0016205578576773405)(106, 0.0016801634337753057)(107, 0.0016243893187493086)(108, 0.0016242009587585926)(109, 0.00174382992554456)(110, 0.0015741429524496198)(111, 0.0016057536704465747)(112, 0.0016985228285193443)(113, 0.0015864471206441522)(114, 0.0015594123397022486)(115, 0.0015755925560370088)(116, 0.0016358800930902362)(117, 0.0015481427544727921)(118, 0.0015468960627913475)(119, 0.0015380959957838058)(120, 0.0015387522289529443)(121, 0.0015802527777850628)(122, 0.0016348466742783785)(123, 0.001564366277307272)(124, 0.0014876658096909523)(125, 0.001563695608638227)(126, 0.0016432928387075663)(127, 0.0015902596060186625)(128, 0.0015060307923704386)(129, 0.0014644560869783163)(130, 0.0014879171503707767)(131, 0.0015073517570272088)(132, 0.0016478761099278927)(133, 0.0015890032518655062)(134, 0.0015514349797740579)(135, 0.001586739206686616)(136, 0.0015005487948656082)(137, 0.0014975822996348143)(138, 0.0015418664552271366)(139, 0.0014961351407691836)(140, 0.0014819980133324862)(141, 0.0015428834594786167)(142, 0.0015886942856013775)(143, 0.0014858978101983666)(144, 0.0014947872841730714)(145, 0.0014997958205640316)(146, 0.001549535314552486)(147, 0.0015244847163558006)(148, 0.001536708907224238)(149, 0.0016246852464973927)(150, 0.0015568407252430916)
					
				};
				\addlegendentry{1 Fixed 0.9}
				
				\addplot[smooth,blue] plot coordinates { 
					(63, 0.02600228600203991)(64, 0.019685635343194008)(65, 0.014955361373722553)(66, 0.013997102156281471)(67, 0.01136621180921793)(68, 0.009515207260847092)(69, 0.008864986710250378)(70, 0.008011426776647568)(71, 0.008077532984316349)(72, 0.0077017247676849365)(73, 0.006835888605564833)(74, 0.0070977285504341125)(75, 0.005754356738179922)(76, 0.005770084448158741)(77, 0.0065780882723629475)(78, 0.00652904761955142)(79, 0.005835945252329111)(80, 0.005592420231550932)(81, 0.006408840883523226)(82, 0.006613102741539478)(83, 0.007747475523501635)(84, 0.008944150060415268)(85, 0.009312134236097336)(86, 0.011422323063015938)(87, 0.01177859678864479)(88, 0.012865141965448856)(89, 0.013734813779592514)(90, 0.012591979466378689)(91, 0.0139646390452981)(92, 0.006821020040661097)(93, 0.004262925125658512)(94, 0.0035977656953036785)(95, 0.0032532205805182457)(96, 0.002930083777755499)(97, 0.0024974560365080833)(98, 0.0023876335471868515)(99, 0.002208967925980687)(100, 0.0024357098154723644)(101, 0.002046189270913601)(102, 0.0020726213697344065)(103, 0.0020001952070742846)(104, 0.0022343932650983334)(105, 0.0020336981397122145)(106, 0.0019368741195648909)(107, 0.0019407692598178983)(108, 0.0020297341980040073)(109, 0.0019186835270375013)(110, 0.001810442190617323)(111, 0.0018387401942163706)(112, 0.0018106253119185567)(113, 0.001721375621855259)(114, 0.0017180790891870856)(115, 0.0017601357540115714)(116, 0.0016651967307552695)(117, 0.0016878234455361962)(118, 0.0016739197308197618)(119, 0.0016121305525302887)(120, 0.00158365024253726)(121, 0.0016692900098860264)(122, 0.00166986882686615)(123, 0.0015743791591376066)(124, 0.0016584604745730758)(125, 0.0015914087416604161)(126, 0.0015154106076806784)(127, 0.001565514481626451)(128, 0.0016196517972275615)(129, 0.001599714858457446)(130, 0.0015511595411226153)(131, 0.0016429191455245018)(132, 0.0015909718349575996)(133, 0.001520646852441132)(134, 0.0016300252173095942)(135, 0.0015867098700255156)(136, 0.001564950100146234)(137, 0.0015831011114642024)(138, 0.0014912193873897195)(139, 0.0017164626624435186)(140, 0.0015989317325875163)(141, 0.0015991732943803072)(142, 0.0015177541645243764)(143, 0.0016000120667740703)(144, 0.0016270738560706377)(145, 0.0016304368618875742)(146, 0.0016333084786310792)(147, 0.0016490442212671041)(148, 0.0017285930225625634)(149, 0.00167648505885154)(150, 0.0016139690997079015)
					
				};
				\addlegendentry{2 Fixed 0.95}
				
				\addplot[smooth,green] plot coordinates { 
					(62, 0.029165469110012054)(63, 0.013110212981700897)(64, 0.009711170569062233)(65, 0.007545338477939367)(66, 0.005954013671725988)(67, 0.0052982233464717865)(68, 0.004900997970253229)(69, 0.004323960281908512)(70, 0.00448256079107523)(71, 0.0037060289178043604)(72, 0.0034796390682458878)(73, 0.003405893687158823)(74, 0.003419856308028102)(75, 0.003236607415601611)(76, 0.003235958982259035)(77, 0.0030538595747202635)(78, 0.0030442385468631983)(79, 0.002902175998315215)(80, 0.003026080084964633)(81, 0.002500619040802121)(82, 0.0027924871537834406)(83, 0.00273570092394948)(84, 0.0035674371756613255)(85, 0.003558562835678458)(86, 0.0038156763184815645)(87, 0.003426407231017947)(88, 0.0038938133511692286)(89, 0.004693416878581047)(90, 0.009345658123493195)(91, 0.007814660668373108)(92, 0.004257595166563988)(93, 0.0033664254005998373)(94, 0.0026515296194702387)(95, 0.002620403654873371)(96, 0.0022675746586173773)(97, 0.0021587093360722065)(98, 0.002108046319335699)(99, 0.0022066952660679817)(100, 0.0018847448518499732)(101, 0.0018965557683259249)(102, 0.002043363405391574)(103, 0.0017944435821846128)(104, 0.001698860782198608)(105, 0.0017478469526395202)(106, 0.0017226714408025146)(107, 0.0016595075139775872)(108, 0.0016374833649024367)(109, 0.0015841476852074265)(110, 0.0016055963933467865)(111, 0.0016234733629971743)(112, 0.0016593157779425383)(113, 0.0015765342395752668)(114, 0.00163640221580863)(115, 0.001697070780210197)(116, 0.001568411709740758)(117, 0.0016824317863211036)(118, 0.0015128415543586016)(119, 0.0016477247700095177)(120, 0.0015508797951042652)(121, 0.0015958822332322598)(122, 0.0015982436016201973)(123, 0.0015331344911828637)(124, 0.001484030857682228)(125, 0.001485506072640419)(126, 0.0015576301375404)(127, 0.0015374536160379648)(128, 0.0015878368867561221)(129, 0.0014873686013743281)(130, 0.0015151652041822672)(131, 0.001413842081092298)(132, 0.001563211902976036)(133, 0.001431287731975317)(134, 0.001466704998165369)(135, 0.001464066095650196)(136, 0.0015122131444513798)(137, 0.001488442881964147)(138, 0.0014440492959693074)(139, 0.001424300018697977)(140, 0.0014794849557802081)(141, 0.0014951586490496993)(142, 0.0014295068103820086)(143, 0.0014853901229798794)(144, 0.0015504173934459686)(145, 0.0014735867734998465)(146, 0.0015837439568713307)(147, 0.0014415105106309056)(148, 0.0015118271112442017)(149, 0.0015411889180541039)(150, 0.0015190194826573133)
					
				};
				\addlegendentry{3 0.95-0.9}	
				
				\addplot[smooth,red] plot coordinates { 
					(62, 0.035874102264642715)(63, 0.016491184011101723)(64, 0.012202675454318523)(65, 0.009570456109941006)(66, 0.008391373790800571)(67, 0.006630809977650642)(68, 0.006224554963409901)(69, 0.005381885450333357)(70, 0.004639514721930027)(71, 0.0041495077311992645)(72, 0.004300115164369345)(73, 0.0038644515443593264)(74, 0.0041479868814349174)(75, 0.0036022483836859465)(76, 0.003568043000996113)(77, 0.003286145394667983)(78, 0.003436790779232979)(79, 0.0030178159940987825)(80, 0.0032151557970792055)(81, 0.0033851834014058113)(82, 0.002869818825274706)(83, 0.0028289388865232468)(84, 0.0029791593551635742)(85, 0.00254795141518116)(86, 0.002983556594699621)(87, 0.0023783135693520308)(88, 0.002579720923677087)(89, 0.0027875045780092478)(90, 0.0028789115604013205)(91, 0.0027010515332221985)(92, 0.002743196440860629)(93, 0.002074351068586111)(94, 0.0021501744631677866)(95, 0.002041538944467902)(96, 0.002115146489813924)(97, 0.0019753912929445505)(98, 0.001964858267456293)(99, 0.0018262859666720033)(100, 0.0018829233013093472)(101, 0.001999726053327322)(102, 0.0017656221752986312)(103, 0.0017836869228631258)(104, 0.0017060467507690191)(105, 0.0018239263445138931)(106, 0.0016824250342324376)(107, 0.001753315911628306)(108, 0.001731721218675375)(109, 0.0016840912867337465)(110, 0.0017039583763107657)(111, 0.0018532072426751256)(112, 0.0016643007984384894)(113, 0.0018274773610755801)(114, 0.0016188591253012419)(115, 0.0017604321474209428)(116, 0.0017485845601186156)(117, 0.0016987572889775038)(118, 0.0017335587181150913)(119, 0.0016411686083301902)(120, 0.0016232117777690291)(121, 0.0016598593210801482)(122, 0.001574530964717269)(123, 0.0017008454306051135)(124, 0.0017246425850316882)(125, 0.0016566970152780414)(126, 0.0016556945629417896)(127, 0.0016322910087183118)(128, 0.001569021842442453)(129, 0.001566341845318675)(130, 0.0016639000969007611)(131, 0.0017213660757988691)(132, 0.0015628081746399403)(133, 0.0016568623250350356)(134, 0.0016858313465490937)(135, 0.0016990738222375512)(136, 0.0015259793726727366)(137, 0.0016708768671378493)(138, 0.001643836498260498)(139, 0.0015557387378066778)(140, 0.0015618576435372233)(141, 0.0015709787840023637)(142, 0.001559063559398055)(143, 0.001587802660651505)(144, 0.001602781587280333)(145, 0.0016273417277261615)(146, 0.0016683819703757763)(147, 0.0015757341170683503)(148, 0.0015737593639642)(149, 0.001664607087150216)(150, 0.0015541431494057178)
					
				};
				\addlegendentry{4 0.9-0.95}
				
			\end{axis}	
		\end{tikzpicture}
		\\Average Loss / Epoch
	\end{minipage}
	\caption{Asymmetric Momentum tested on WRN28-10 with Cifar10. The fourth group with the asymmetric momentum achieved the best performance. At the same time, we can clearly observe that after swapping the momentum direction, the acceleration momentum in green exhibits the same missing optimal destination. However, the red does not show this phenomenon at all. This confirms that weights have a very distinct direction specificity, and the missing is primarily caused by momentum towards $\Theta_s$, it also means Cifar10 is a kind of $\Theta_n$ lead non-sparse datasets. Note that this is not the final result, but rather a rapid reduction in the learning rate to test the ability to overcome saddle points.}
	\label{Fig0401}
\end{figure*}

In this experimental setup in Figure.\ref{Fig0401}, the impact brought by the distinct direction specificity is very evident.

We can observe that the second group, which maintains a momentum of 0.95 throughout, has an excessive momentum that prevents it from making timely turns before reaching the destination, thus missing the optimal point. Compared to the first group, there is a significant abnormal increase in both its error rate and loss value.

In the third group, we observed a similar phenomenon. Compared to the fixed 0.95 momentum, its abnormal increase occurred later. This is because, even though we were continuously pushing to the right, or the sparse direction, the bias should be more. However, due to the speed of descending along the gradient descent, which is slower compared to the continuous 0.95 momentum in the second group, it caused the timing of the abnormality to be later than the fixed 0.95 momentum group.

Looking further into the fourth group, \textbf{only swap momentum directions} in previous group, which uses a momentum of 0.95 during the non-sparse phase, it's evident that no anomalies occurred, and its final performance is significantly better than the other groups. This clearly demonstrates that the properties of $\Theta_n$ and $\Theta_s$ are distinct, indicating that the weights are specificity in nature. Furthermore, it also shows that Cifar10 is a non-sparse dataset, consistent with the composition of datasets.

\subsection{Cifar100}

In the Cifar100 dataset, the situation underwent a noticeable change. Even though we set up the same experimental combinations, the results were distinctly different from those in Cifar10. The four groups are same as Cifar10.
%
%
%
%
%

\begin{figure*}[h]
	\centering
	\begin{minipage}[h]{0.4\textwidth}
		\centering
		\begin{tikzpicture} 
			\begin{axis}[
				scale=0.6,
				tick align=inside, 
				xtick = {0,50,100,150},
				ytick = {20,40,60},
				grid style={dashed},
				xmajorgrids =true,
				ymajorgrids =true,
				legend style={at={(1,0.8)},anchor=east} 
				]
				
					%
				\addplot[smooth,black] plot coordinates { 
					(4, 61.459999084472656)(5, 56.060001373291016)(6, 51.02000045776367)(7, 51.04999923706055)(8, 52.900001525878906)(9, 49.09000015258789)(10, 50.130001068115234)(11, 43.41999816894531)(12, 48.060001373291016)(13, 51.06999969482422)(14, 44.18000030517578)(15, 45.849998474121094)(16, 46.060001373291016)(17, 44.5099983215332)(18, 44.86000061035156)(19, 44.220001220703125)(20, 44.93000030517578)(21, 40.630001068115234)(22, 43.400001525878906)(23, 42.52000045776367)(24, 43.900001525878906)(25, 40.369998931884766)(26, 42.97999954223633)(27, 41.52000045776367)(28, 44.25)(29, 40.189998626708984)(30, 43.04999923706055)(31, 41.599998474121094)(32, 23.329999923706055)(33, 23.450000762939453)(34, 23.5)(35, 23.729999542236328)(36, 24.309999465942383)(37, 24.309999465942383)(38, 25.809999465942383)(39, 26.479999542236328)(40, 26.479999542236328)(41, 26.770000457763672)(42, 26.8700008392334)(43, 28.3799991607666)(44, 29.190000534057617)(45, 28.469999313354492)(46, 29.200000762939453)(47, 29.149999618530273)(48, 29.299999237060547)(49, 29.93000030517578)(50, 28.010000228881836)(51, 29.350000381469727)(52, 31.610000610351562)(53, 28.81999969482422)(54, 29.809999465942383)(55, 28.34000015258789)(56, 29.690000534057617)(57, 30.049999237060547)(58, 29.649999618530273)(59, 31.219999313354492)(60, 29.34000015258789)(61, 31.3700008392334)(62, 21.31999969482422)(63, 21.31999969482422)(64, 20.979999542236328)(65, 21.030000686645508)(66, 20.8700008392334)(67, 20.860000610351562)(68, 20.670000076293945)(69, 20.690000534057617)(70, 20.739999771118164)(71, 20.690000534057617)(72, 20.780000686645508)(73, 20.829999923706055)(74, 20.770000457763672)(75, 20.690000534057617)(76, 20.649999618530273)(77, 20.829999923706055)(78, 20.690000534057617)(79, 20.610000610351562)(80, 20.25)(81, 20.510000228881836)(82, 20.399999618530273)(83, 20.420000076293945)(84, 20.520000457763672)(85, 20.549999237060547)(86, 20.34000015258789)(87, 20.420000076293945)(88, 20.299999237060547)(89, 20.290000915527344)(90, 20.200000762939453)(91, 20.059999465942383)(92, 20.209999084472656)(93, 20.270000457763672)(94, 20.219999313354492)(95, 20.200000762939453)(96, 20.290000915527344)(97, 20.190000534057617)(98, 20.350000381469727)(99, 20.200000762939453)(100, 20.170000076293945)(101, 20.170000076293945)(102, 20.190000534057617)(103, 20.1299991607666)(104, 20.139999389648438)(105, 20.1200008392334)(106, 20.389999389648438)(107, 20.030000686645508)(108, 19.989999771118164)(109, 20.15999984741211)(110, 20.09000015258789)(111, 20.06999969482422)(112, 20.329999923706055)(113, 20.280000686645508)(114, 20.360000610351562)(115, 20.1299991607666)(116, 20.059999465942383)(117, 20.360000610351562)(118, 20.219999313354492)(119, 20.139999389648438)(120, 20.149999618530273)(121, 20.100000381469727)(122, 20.100000381469727)(123, 20.040000915527344)(124, 20.110000610351562)(125, 20.110000610351562)(126, 20.09000015258789)(127, 19.950000762939453)(128, 20.049999237060547)(129, 20.18000030517578)(130, 20.030000686645508)(131, 20.059999465942383)(132, 20.209999084472656)(133, 20.100000381469727)(134, 20.059999465942383)(135, 20.219999313354492)(136, 20.06999969482422)(137, 20.049999237060547)(138, 20.079999923706055)(139, 20.049999237060547)(140, 20.260000228881836)(141, 20.030000686645508)(142, 20.219999313354492)(143, 20.059999465942383)(144, 19.93000030517578)(145, 20.1200008392334)(146, 20.049999237060547)(147, 20.200000762939453)(148, 20.139999389648438)(149, 20.15999984741211)(150, 19.959999084472656)
					
				};
				
				\addplot[smooth,blue] plot coordinates { 
					(5, 60.9900016784668)(6, 61.2400016784668)(7, 62.90999984741211)(8, 59.43000030517578)(9, 55.65999984741211)(10, 60.290000915527344)(11, 53.029998779296875)(12, 52.540000915527344)(13, 54.84000015258789)(14, 54.560001373291016)(15, 52.650001525878906)(16, 52.97999954223633)(17, 52.15999984741211)(18, 50.0099983215332)(19, 43.959999084472656)(20, 45.88999938964844)(21, 47.88999938964844)(22, 48.68000030517578)(23, 50.16999816894531)(24, 52.970001220703125)(25, 47.18000030517578)(26, 51.59000015258789)(27, 51.880001068115234)(28, 46.970001220703125)(29, 54.790000915527344)(30, 48.810001373291016)(31, 51.900001525878906)(32, 28.25)(33, 27.1200008392334)(34, 28.8700008392334)(35, 28.860000610351562)(36, 28.3700008392334)(37, 30.510000228881836)(38, 31.040000915527344)(39, 30.520000457763672)(40, 31.780000686645508)(41, 31.079999923706055)(42, 31.149999618530273)(43, 32.4900016784668)(44, 32.77000045776367)(45, 30.850000381469727)(46, 31.440000534057617)(47, 31.229999542236328)(48, 31.719999313354492)(49, 30.59000015258789)(50, 32.29999923706055)(51, 31.040000915527344)(52, 29.600000381469727)(53, 31.200000762939453)(54, 30.950000762939453)(55, 31.040000915527344)(56, 30.450000762939453)(57, 30.989999771118164)(58, 33.470001220703125)(59, 30.90999984741211)(60, 32.27000045776367)(61, 32.18000030517578)(62, 22.06999969482422)(63, 21.440000534057617)(64, 21.25)(65, 21.440000534057617)(66, 21.440000534057617)(67, 21.280000686645508)(68, 21.110000610351562)(69, 21.280000686645508)(70, 21.06999969482422)(71, 20.950000762939453)(72, 21.09000015258789)(73, 21.170000076293945)(74, 20.969999313354492)(75, 21.030000686645508)(76, 20.850000381469727)(77, 20.940000534057617)(78, 21.079999923706055)(79, 21.09000015258789)(80, 21.010000228881836)(81, 21.270000457763672)(82, 20.93000030517578)(83, 20.420000076293945)(84, 20.520000457763672)(85, 20.950000762939453)(86, 21.229999542236328)(87, 21.40999984741211)(88, 21.1299991607666)(89, 21.010000228881836)(90, 20.93000030517578)(91, 21.110000610351562)(92, 20.65999984741211)(93, 20.540000915527344)(94, 20.399999618530273)(95, 20.520000457763672)(96, 20.479999542236328)(97, 20.450000762939453)(98, 20.729999542236328)(99, 20.399999618530273)(100, 20.59000015258789)(101, 20.479999542236328)(102, 20.510000228881836)(103, 20.610000610351562)(104, 20.530000686645508)(105, 20.520000457763672)(106, 20.610000610351562)(107, 20.530000686645508)(108, 20.700000762939453)(109, 20.360000610351562)(110, 20.469999313354492)(111, 20.549999237060547)(112, 20.600000381469727)(113, 20.6200008392334)(114, 20.549999237060547)(115, 20.729999542236328)(116, 20.649999618530273)(117, 20.760000228881836)(118, 20.639999389648438)(119, 20.540000915527344)(120, 20.540000915527344)(121, 20.579999923706055)(122, 20.530000686645508)(123, 20.719999313354492)(124, 20.520000457763672)(125, 20.709999084472656)(126, 20.6200008392334)(127, 20.559999465942383)(128, 20.75)(129, 20.59000015258789)(130, 20.459999084472656)(131, 20.799999237060547)(132, 20.709999084472656)(133, 20.520000457763672)(134, 20.56999969482422)(135, 20.549999237060547)(136, 20.639999389648438)(137, 20.780000686645508)(138, 20.639999389648438)(139, 20.610000610351562)(140, 20.770000457763672)(141, 20.709999084472656)(142, 20.579999923706055)(143, 20.8799991607666)(144, 20.510000228881836)(145, 20.530000686645508)(146, 20.709999084472656)(147, 20.739999771118164)(148, 20.540000915527344)(149, 20.760000228881836)(150, 20.760000228881836)
					
				};
				
				\addplot[smooth,green] plot coordinates { 
					(7, 52.27000045776367)(8, 58.06999969482422)(9, 50.709999084472656)(10, 50.599998474121094)(11, 48.66999816894531)(12, 55.279998779296875)(13, 48.150001525878906)(14, 47.16999816894531)(15, 45.470001220703125)(16, 45.439998626708984)(17, 48.790000915527344)(18, 48.83000183105469)(19, 45.599998474121094)(20, 41.79999923706055)(21, 42.68000030517578)(22, 41.79999923706055)(23, 42.220001220703125)(24, 43.5)(25, 41.119998931884766)(26, 39.900001525878906)(27, 41.45000076293945)(28, 45.040000915527344)(29, 41.40999984741211)(30, 45.79999923706055)(31, 43.06999969482422)(32, 25.3700008392334)(33, 24.190000534057617)(34, 24.989999771118164)(35, 25.170000076293945)(36, 26.209999084472656)(37, 26.799999237060547)(38, 26.979999542236328)(39, 27.59000015258789)(40, 28.270000457763672)(41, 27.799999237060547)(42, 28.100000381469727)(43, 27.100000381469727)(44, 29.139999389648438)(45, 28.65999984741211)(46, 27.84000015258789)(47, 28.020000457763672)(48, 27.889999389648438)(49, 29.8700008392334)(50, 30.309999465942383)(51, 29.6299991607666)(52, 27.950000762939453)(53, 29.90999984741211)(54, 28.229999542236328)(55, 29.549999237060547)(56, 31.979999542236328)(57, 30.299999237060547)(58, 29.770000457763672)(59, 29.510000228881836)(60, 29.270000457763672)(61, 29.75)(62, 21.8799991607666)(63, 20.969999313354492)(64, 20.860000610351562)(65, 20.899999618530273)(66, 20.59000015258789)(67, 20.68000030517578)(68, 20.700000762939453)(69, 20.690000534057617)(70, 20.450000762939453)(71, 20.239999771118164)(72, 20.469999313354492)(73, 20.15999984741211)(74, 20.190000534057617)(75, 20.450000762939453)(76, 20.059999465942383)(77, 20.149999618530273)(78, 20.219999313354492)(79, 20.190000534057617)(80, 20.0)(81, 19.989999771118164)(82, 19.739999771118164)(83, 20.100000381469727)(84, 20.270000457763672)(85, 20.280000686645508)(86, 20.15999984741211)(87, 20.040000915527344)(88, 20.139999389648438)(89, 20.09000015258789)(90, 19.899999618530273)(91, 20.040000915527344)(92, 20.010000228881836)(93, 19.84000015258789)(94, 19.739999771118164)(95, 19.860000610351562)(96, 19.84000015258789)(97, 19.829999923706055)(98, 19.709999084472656)(99, 19.729999542236328)(100, 19.940000534057617)(101, 19.809999465942383)(102, 19.829999923706055)(103, 19.690000534057617)(104, 19.75)(105, 20.059999465942383)(106, 19.93000030517578)(107, 19.739999771118164)(108, 19.8700008392334)(109, 19.899999618530273)(110, 19.8700008392334)(111, 19.780000686645508)(112, 19.81999969482422)(113, 19.8799991607666)(114, 19.760000228881836)(115, 19.59000015258789)(116, 19.760000228881836)(117, 19.719999313354492)(118, 19.799999237060547)(119, 19.84000015258789)(120, 19.889999389648438)(121, 19.700000762939453)(122, 19.889999389648438)(123, 19.739999771118164)(124, 20.049999237060547)(125, 19.600000381469727)(126, 19.809999465942383)(127, 19.770000457763672)(128, 19.979999542236328)(129, 19.729999542236328)(130, 19.760000228881836)(131, 19.850000381469727)(132, 19.989999771118164)(133, 19.84000015258789)(134, 19.860000610351562)(135, 19.790000915527344)(136, 19.940000534057617)(137, 20.0)(138, 19.959999084472656)(139, 19.739999771118164)(140, 19.709999084472656)(141, 19.739999771118164)(142, 19.780000686645508)(143, 19.850000381469727)(144, 19.8700008392334)(145, 19.739999771118164)(146, 19.6299991607666)(147, 19.690000534057617)(148, 19.90999984741211)(149, 19.760000228881836)(150, 19.799999237060547)

				};
				
				\addplot[smooth,red] plot coordinates { 
					(5, 56.040000915527344)(6, 54.189998626708984)(7, 58.650001525878906)(8, 59.70000076293945)(9, 55.959999084472656)(10, 48.13999938964844)(11, 44.849998474121094)(12, 43.970001220703125)(13, 41.959999084472656)(14, 48.7599983215332)(15, 45.13999938964844)(16, 42.0099983215332)(17, 42.68000030517578)(18, 41.36000061035156)(19, 43.0)(20, 52.630001068115234)(21, 44.0)(22, 37.459999084472656)(23, 38.959999084472656)(24, 42.439998626708984)(25, 39.09000015258789)(26, 39.15999984741211)(27, 42.779998779296875)(28, 38.72999954223633)(29, 38.369998931884766)(30, 36.279998779296875)(31, 38.13999938964844)(32, 26.100000381469727)(33, 25.270000457763672)(34, 26.059999465942383)(35, 27.520000457763672)(36, 28.729999542236328)(37, 29.639999389648438)(38, 31.459999084472656)(39, 32.0099983215332)(40, 32.27000045776367)(41, 32.369998931884766)(42, 30.020000457763672)(43, 37.0)(44, 33.599998474121094)(45, 32.540000915527344)(46, 34.77000045776367)(47, 28.75)(48, 33.959999084472656)(49, 26.270000457763672)(50, 36.790000915527344)(51, 27.670000076293945)(52, 33.56999969482422)(53, 27.059999465942383)(54, 35.68000030517578)(55, 24.309999465942383)(56, 35.060001373291016)(57, 26.90999984741211)(58, 34.290000915527344)(59, 25.579999923706055)(60, 36.220001220703125)(61, 25.950000762939453)(62, 21.309999465942383)(63, 21.280000686645508)(64, 21.209999084472656)(65, 20.969999313354492)(66, 21.1200008392334)(67, 21.06999969482422)(68, 21.030000686645508)(69, 20.81999969482422)(70, 20.790000915527344)(71, 20.639999389648438)(72, 20.75)(73, 20.670000076293945)(74, 20.549999237060547)(75, 20.68000030517578)(76, 20.510000228881836)(77, 20.770000457763672)(78, 20.90999984741211)(79, 20.5)(80, 20.479999542236328)(81, 20.510000228881836)(82, 20.780000686645508)(83, 20.770000457763672)(84, 20.799999237060547)(85, 20.6299991607666)(86, 20.850000381469727)(87, 20.8799991607666)(88, 20.739999771118164)(89, 20.440000534057617)(90, 20.5)(91, 20.670000076293945)(92, 20.56999969482422)(93, 20.56999969482422)(94, 20.649999618530273)(95, 20.719999313354492)(96, 20.68000030517578)(97, 20.559999465942383)(98, 20.489999771118164)(99, 20.75)(100, 20.579999923706055)(101, 20.68000030517578)(102, 20.790000915527344)(103, 20.65999984741211)(104, 20.579999923706055)(105, 20.729999542236328)(106, 20.549999237060547)(107, 20.670000076293945)(108, 20.43000030517578)(109, 20.489999771118164)(110, 20.559999465942383)(111, 20.479999542236328)(112, 20.559999465942383)(113, 20.700000762939453)(114, 20.610000610351562)(115, 20.610000610351562)(116, 20.639999389648438)(117, 20.520000457763672)(118, 20.43000030517578)(119, 20.520000457763672)(120, 20.450000762939453)(121, 20.420000076293945)(122, 20.559999465942383)(123, 20.549999237060547)(124, 20.489999771118164)(125, 20.329999923706055)(126, 20.56999969482422)(127, 20.600000381469727)(128, 20.739999771118164)(129, 20.579999923706055)(130, 20.559999465942383)(131, 20.459999084472656)(132, 20.639999389648438)(133, 20.639999389648438)(134, 20.600000381469727)(135, 20.31999969482422)(136, 20.530000686645508)(137, 20.559999465942383)(138, 20.65999984741211)(139, 20.6200008392334)(140, 20.65999984741211)(141, 20.549999237060547)(142, 20.520000457763672)(143, 20.489999771118164)(144, 20.579999923706055)(145, 20.5)(146, 20.530000686645508)(147, 20.440000534057617)(148, 20.459999084472656)(149, 20.329999923706055)(150, 20.399999618530273)
					
				};

			\end{axis}	
		\end{tikzpicture}
		\\Test Error / Epoch
	\end{minipage}\begin{minipage}[h]{0.4\textwidth}
		\centering
		\begin{tikzpicture} 
			\begin{axis}[
				scale=0.6,
				tick align=inside, 
				ymode=log,
				grid style={dashed},
				xmajorgrids =true,
				ymajorgrids =true,
				legend style={at={(1.4,1)},anchor=north} 
				]
				
					%
				\addplot[smooth,black] plot coordinates { 
					(1, 3.909233331680298)(2, 3.2083370685577393)(3, 2.6051745414733887)(4, 2.177917003631592)(5, 1.9037399291992188)(6, 1.7190378904342651)(7, 1.588405728340149)(8, 1.4909778833389282)(9, 1.406525731086731)(10, 1.345362901687622)(11, 1.2975462675094604)(12, 1.2512561082839966)(13, 1.2127552032470703)(14, 1.183013916015625)(15, 1.1414562463760376)(16, 1.112154483795166)(17, 1.0967912673950195)(18, 1.0546929836273193)(19, 1.0463517904281616)(20, 1.0351898670196533)(21, 1.0144813060760498)(22, 1.0000112056732178)(23, 0.992870569229126)(24, 0.9707098603248596)(25, 0.9675177931785583)(26, 0.9525606036186218)(27, 0.942079484462738)(28, 0.9303240180015564)(29, 0.9223703145980835)(30, 0.9204598665237427)(31, 0.9033728837966919)(32, 0.44860467314720154)(33, 0.30852431058883667)(34, 0.25050830841064453)(35, 0.21260519325733185)(36, 0.1843181699514389)(37, 0.16686974465847015)(38, 0.15439197421073914)(39, 0.15496034920215607)(40, 0.14895452558994293)(41, 0.14784805476665497)(42, 0.16239549219608307)(43, 0.15981806814670563)(44, 0.16754736006259918)(45, 0.17148812115192413)(46, 0.16861741244792938)(47, 0.18210674822330475)(48, 0.17778143286705017)(49, 0.19597287476062775)(50, 0.17807702720165253)(51, 0.1862885057926178)(52, 0.1797400414943695)(53, 0.18158484995365143)(54, 0.1775851547718048)(55, 0.17114943265914917)(56, 0.16978557407855988)(57, 0.1710764616727829)(58, 0.16903547942638397)(59, 0.16566254198551178)(60, 0.1681661605834961)(61, 0.15372906625270844)(62, 0.05073586851358414)(63, 0.02132103405892849)(64, 0.01537653710693121)(65, 0.013287446461617947)(66, 0.011323802173137665)(67, 0.010326031595468521)(68, 0.009749882854521275)(69, 0.009338966570794582)(70, 0.00831600371748209)(71, 0.008183675818145275)(72, 0.007893932051956654)(73, 0.00779728451743722)(74, 0.007601246703416109)(75, 0.007879339158535004)(76, 0.007543621119111776)(77, 0.007252804934978485)(78, 0.007596908137202263)(79, 0.007442531641572714)(80, 0.007326944265514612)(81, 0.007060553878545761)(82, 0.007325255312025547)(83, 0.007680289447307587)(84, 0.00738784484565258)(85, 0.007548435125499964)(86, 0.007224962115287781)(87, 0.007582414895296097)(88, 0.00773333664983511)(89, 0.007585266139358282)(90, 0.007298631593585014)(91, 0.007545977830886841)(92, 0.0073680817149579525)(93, 0.007098803762346506)(94, 0.006848534103482962)(95, 0.006924607325345278)(96, 0.006784286815673113)(97, 0.006838986650109291)(98, 0.006916705518960953)(99, 0.00676843011751771)(100, 0.006738549564033747)(101, 0.007044902537018061)(102, 0.006849419791251421)(103, 0.006800173316150904)(104, 0.006730925291776657)(105, 0.0067961071617901325)(106, 0.006802281364798546)(107, 0.006652679294347763)(108, 0.006688783876597881)(109, 0.00673556188121438)(110, 0.006716983858495951)(111, 0.006788548082113266)(112, 0.006755623035132885)(113, 0.006922314874827862)(114, 0.006684582680463791)(115, 0.00683830538764596)(116, 0.006596560124307871)(117, 0.006924137007445097)(118, 0.006953797303140163)(119, 0.00664180563762784)(120, 0.007016538176685572)(121, 0.006680131424218416)(122, 0.006839394103735685)(123, 0.0066690449602901936)(124, 0.0067183091305196285)(125, 0.006845915224403143)(126, 0.006979095749557018)(127, 0.006853064987808466)(128, 0.006802326068282127)(129, 0.006847653072327375)(130, 0.006919819861650467)(131, 0.006969415117055178)(132, 0.006835187319666147)(133, 0.006724765989929438)(134, 0.0066922507248818874)(135, 0.007042733486741781)(136, 0.007034815847873688)(137, 0.006791203282773495)(138, 0.006971009541302919)(139, 0.006858963053673506)(140, 0.006976872682571411)(141, 0.007035657297819853)(142, 0.006815703120082617)(143, 0.006970244459807873)(144, 0.006999331526458263)(145, 0.00694572227075696)(146, 0.007133933249861002)(147, 0.007096423767507076)(148, 0.00705445371568203)(149, 0.00695949187502265)(150, 0.007177927531301975)
					
				};
				\addlegendentry{1 Fixed 0.9}
				
				\addplot[smooth,blue] plot coordinates { 
					(1, 4.058912754058838)(2, 3.579803705215454)(3, 3.10744309425354)(4, 2.594338893890381)(5, 2.2472896575927734)(6, 2.012104034423828)(7, 1.8628771305084229)(8, 1.7751151323318481)(9, 1.676296353340149)(10, 1.6222155094146729)(11, 1.5782155990600586)(12, 1.5430995225906372)(13, 1.509371280670166)(14, 1.488109827041626)(15, 1.4566011428833008)(16, 1.4460656642913818)(17, 1.4226622581481934)(18, 1.421347737312317)(19, 1.4130072593688965)(20, 1.3976514339447021)(21, 1.3903400897979736)(22, 1.3810192346572876)(23, 1.371301293373108)(24, 1.3585244417190552)(25, 1.3558610677719116)(26, 1.3525567054748535)(27, 1.3299819231033325)(28, 1.3381705284118652)(29, 1.3366289138793945)(30, 1.3280531167984009)(31, 1.3325992822647095)(32, 0.8206137418746948)(33, 0.6539905667304993)(34, 0.5928373336791992)(35, 0.5472848415374756)(36, 0.5272560119628906)(37, 0.5135694742202759)(38, 0.5089945793151855)(39, 0.5082470178604126)(40, 0.4943862557411194)(41, 0.4977724254131317)(42, 0.48470085859298706)(43, 0.4847339391708374)(44, 0.4540863633155823)(45, 0.44772472977638245)(46, 0.4375315010547638)(47, 0.44137686491012573)(48, 0.42192745208740234)(49, 0.41280966997146606)(50, 0.4016624093055725)(51, 0.4053858816623688)(52, 0.39411741495132446)(53, 0.39163529872894287)(54, 0.3796791732311249)(55, 0.3685349225997925)(56, 0.36822718381881714)(57, 0.36301419138908386)(58, 0.36114370822906494)(59, 0.36908969283103943)(60, 0.35212698578834534)(61, 0.36130157113075256)(62, 0.1439915895462036)(63, 0.07106659561395645)(64, 0.05320350453257561)(65, 0.0442926362156868)(66, 0.03782567381858826)(67, 0.0339193269610405)(68, 0.029734386131167412)(69, 0.026971060782670975)(70, 0.026122136041522026)(71, 0.024738358333706856)(72, 0.023388635367155075)(73, 0.022190669551491737)(74, 0.021587802097201347)(75, 0.02058470994234085)(76, 0.0207288209348917)(77, 0.01902065984904766)(78, 0.01936877891421318)(79, 0.017964214086532593)(80, 0.01853233203291893)(81, 0.018734268844127655)(82, 0.01809949427843094)(83, 0.01803344115614891)(84, 0.017942966893315315)(85, 0.01779882237315178)(86, 0.016949763521552086)(87, 0.017318785190582275)(88, 0.016596775501966476)(89, 0.018717866390943527)(90, 0.016430338844656944)(91, 0.017136840149760246)(92, 0.014438183978199959)(93, 0.01305594015866518)(94, 0.012068761512637138)(95, 0.011739044450223446)(96, 0.011894283816218376)(97, 0.01174051035195589)(98, 0.011452321894466877)(99, 0.011253527365624905)(100, 0.011054362170398235)(101, 0.011322845704853535)(102, 0.011016178876161575)(103, 0.010899248532950878)(104, 0.011298920027911663)(105, 0.010958598926663399)(106, 0.010864941403269768)(107, 0.010789061896502972)(108, 0.010703930631279945)(109, 0.01080623921006918)(110, 0.01092724408954382)(111, 0.010653920471668243)(112, 0.01069159060716629)(113, 0.010732650756835938)(114, 0.010473188012838364)(115, 0.01054544746875763)(116, 0.010714096017181873)(117, 0.010544061660766602)(118, 0.010607743635773659)(119, 0.01056178379803896)(120, 0.01048623863607645)(121, 0.010717080906033516)(122, 0.01048739068210125)(123, 0.01042154524475336)(124, 0.01040159072726965)(125, 0.01036912202835083)(126, 0.01048220507800579)(127, 0.01036909967660904)(128, 0.010658485814929008)(129, 0.010416113771498203)(130, 0.01052214577794075)(131, 0.010163231752812862)(132, 0.010388717986643314)(133, 0.010642820037901402)(134, 0.010503838770091534)(135, 0.010246802121400833)(136, 0.01038255076855421)(137, 0.010318046435713768)(138, 0.01033033523708582)(139, 0.010662016458809376)(140, 0.010178936645388603)(141, 0.01016808021813631)(142, 0.010368027724325657)(143, 0.010237778536975384)(144, 0.01025388389825821)(145, 0.010077555663883686)(146, 0.010227469727396965)(147, 0.009948169812560081)(148, 0.010106083005666733)(149, 0.0101177291944623)(150, 0.010106083005666733)
					
				};
				\addlegendentry{2 Fixed 0.95}
				
				\addplot[smooth,green] plot coordinates { 
					(1, 4.000728607177734)(2, 3.478665828704834)(3, 2.9675190448760986)(4, 2.4497222900390625)(5, 2.0917563438415527)(6, 1.88906991481781)(7, 1.724629521369934)(8, 1.5933576822280884)(9, 1.525797724723816)(10, 1.4559314250946045)(11, 1.3940037488937378)(12, 1.3313299417495728)(13, 1.2989498376846313)(14, 1.259287714958191)(15, 1.229792594909668)(16, 1.196102499961853)(17, 1.2030270099639893)(18, 1.1800215244293213)(19, 1.1318708658218384)(20, 1.1350951194763184)(21, 1.1134480237960815)(22, 1.1106775999069214)(23, 1.0893908739089966)(24, 1.0692932605743408)(25, 1.081221580505371)(26, 1.0231297016143799)(27, 1.0652761459350586)(28, 1.049421787261963)(29, 1.0433895587921143)(30, 1.0325807332992554)(31, 1.017054557800293)(32, 0.5569714903831482)(33, 0.39160341024398804)(34, 0.3253374993801117)(35, 0.2859852910041809)(36, 0.2511816620826721)(37, 0.24582694470882416)(38, 0.24878287315368652)(39, 0.23193328082561493)(40, 0.230791836977005)(41, 0.2326458990573883)(42, 0.227341890335083)(43, 0.22501054406166077)(44, 0.2270238697528839)(45, 0.23601692914962769)(46, 0.2278878092765808)(47, 0.2175394743680954)(48, 0.20958860218524933)(49, 0.2338920682668686)(50, 0.22822262346744537)(51, 0.23283812403678894)(52, 0.22433575987815857)(53, 0.2042207419872284)(54, 0.2167375385761261)(55, 0.2103252410888672)(56, 0.22641758620738983)(57, 0.23370614647865295)(58, 0.2274290770292282)(59, 0.19420252740383148)(60, 0.19547395408153534)(61, 0.19439715147018433)(62, 0.05845492333173752)(63, 0.023602889850735664)(64, 0.01854417845606804)(65, 0.015851350501179695)(66, 0.01312828715890646)(67, 0.012662091292440891)(68, 0.011200257577002048)(69, 0.010407106950879097)(70, 0.010257705114781857)(71, 0.010011143982410431)(72, 0.009746488183736801)(73, 0.009396979585289955)(74, 0.00894012488424778)(75, 0.009102456271648407)(76, 0.009253890253603458)(77, 0.009159022010862827)(78, 0.009552828036248684)(79, 0.009263643063604832)(80, 0.009384849108755589)(81, 0.009086625650525093)(82, 0.009305810555815697)(83, 0.009531024843454361)(84, 0.009278652258217335)(85, 0.009686943143606186)(86, 0.009153367020189762)(87, 0.009246235713362694)(88, 0.00973949022591114)(89, 0.009713309817016125)(90, 0.009660841897130013)(91, 0.010359051637351513)(92, 0.008911960758268833)(93, 0.008653064258396626)(94, 0.008241540752351284)(95, 0.008195635862648487)(96, 0.008274604566395283)(97, 0.008085806854069233)(98, 0.007774148602038622)(99, 0.007642720825970173)(100, 0.007806041277945042)(101, 0.007855812087655067)(102, 0.007678481284528971)(103, 0.007785685360431671)(104, 0.007587619125843048)(105, 0.007470463868230581)(106, 0.00763496570289135)(107, 0.00767728453502059)(108, 0.007421867921948433)(109, 0.00775828817859292)(110, 0.007868013344705105)(111, 0.007803729735314846)(112, 0.007774568162858486)(113, 0.007746331859380007)(114, 0.007590054534375668)(115, 0.0076913912780582905)(116, 0.007718780543655157)(117, 0.007469248957931995)(118, 0.007703982759267092)(119, 0.007506578229367733)(120, 0.007590699940919876)(121, 0.007542877923697233)(122, 0.007793470751494169)(123, 0.007772678043693304)(124, 0.0075322301127016544)(125, 0.0075499992817640305)(126, 0.007533208467066288)(127, 0.007512545213103294)(128, 0.007602573838084936)(129, 0.007597201503813267)(130, 0.007882206700742245)(131, 0.007645495235919952)(132, 0.007628684863448143)(133, 0.007676086854189634)(134, 0.0075467173010110855)(135, 0.007791782263666391)(136, 0.007567553780972958)(137, 0.0076991659589111805)(138, 0.007577313110232353)(139, 0.007600804790854454)(140, 0.007554531563073397)(141, 0.0075310091488063335)(142, 0.0075588044710457325)(143, 0.007593621499836445)(144, 0.007619835436344147)(145, 0.007515590172261)(146, 0.007485049776732922)(147, 0.007674310822039843)(148, 0.007451333105564117)(149, 0.007507777772843838)(150, 0.0076736584305763245)
					
				};
				\addlegendentry{3 0.95-0.9}
				
				\addplot[smooth,red] plot coordinates { 
					(1, 3.950348138809204)(2, 3.2815184593200684)(3, 2.6591272354125977)(4, 2.2432010173797607)(5, 1.974531888961792)(6, 1.795928716659546)(7, 1.6923024654388428)(8, 1.5962865352630615)(9, 1.4999761581420898)(10, 1.4572690725326538)(11, 1.4276950359344482)(12, 1.4064065217971802)(13, 1.3656237125396729)(14, 1.3766021728515625)(15, 1.3331403732299805)(16, 1.2899807691574097)(17, 1.2856322526931763)(18, 1.2718393802642822)(19, 1.2808862924575806)(20, 1.1940020322799683)(21, 1.3015937805175781)(22, 1.2575421333312988)(23, 1.1745176315307617)(24, 1.2499096393585205)(25, 1.1809500455856323)(26, 1.2491425275802612)(27, 1.1133391857147217)(28, 1.2381724119186401)(29, 1.1590735912322998)(30, 1.2007286548614502)(31, 1.182224988937378)(32, 0.6359008550643921)(33, 0.5249930620193481)(34, 0.4579557180404663)(35, 0.41897061467170715)(36, 0.38388463854789734)(37, 0.3743574321269989)(38, 0.3650747835636139)(39, 0.36693528294563293)(40, 0.33074018359184265)(41, 0.3504895567893982)(42, 0.2957131266593933)(43, 0.33057066798210144)(44, 0.3219365179538727)(45, 0.27198269963264465)(46, 0.3123147487640381)(47, 0.27349987626075745)(48, 0.4225737750530243)(49, 0.26984483003616333)(50, 0.4058693051338196)(51, 0.24925383925437927)(52, 0.4147678315639496)(53, 0.2484448403120041)(54, 0.36124512553215027)(55, 0.21706639230251312)(56, 0.36164650321006775)(57, 0.21953940391540527)(58, 0.26740097999572754)(59, 0.20192615687847137)(60, 0.3703403174877167)(61, 0.210318922996521)(62, 0.06705418229103088)(63, 0.04102883115410805)(64, 0.03162755072116852)(65, 0.026423025876283646)(66, 0.024135952815413475)(67, 0.02035258337855339)(68, 0.01959466189146042)(69, 0.01904863864183426)(70, 0.017302796244621277)(71, 0.016576159745454788)(72, 0.015359400771558285)(73, 0.014689834788441658)(74, 0.014731998555362225)(75, 0.014303241856396198)(76, 0.013822829350829124)(77, 0.012518571689724922)(78, 0.012414820492267609)(79, 0.012000194750726223)(80, 0.012341058813035488)(81, 0.011914005503058434)(82, 0.011689024977385998)(83, 0.011908446438610554)(84, 0.011596220545470715)(85, 0.012252524495124817)(86, 0.01161755621433258)(87, 0.012164300307631493)(88, 0.01113693043589592)(89, 0.011202975176274776)(90, 0.011066884733736515)(91, 0.011183077469468117)(92, 0.009918455965816975)(93, 0.009713614359498024)(94, 0.009794745594263077)(95, 0.009235205128788948)(96, 0.009101598523557186)(97, 0.009178709238767624)(98, 0.009263645857572556)(99, 0.008971945382654667)(100, 0.008875604718923569)(101, 0.009072872810065746)(102, 0.008800298906862736)(103, 0.009030400775372982)(104, 0.008695077151060104)(105, 0.008747624233365059)(106, 0.008637405931949615)(107, 0.008726474829018116)(108, 0.008718996308743954)(109, 0.008958484046161175)(110, 0.00884394720196724)(111, 0.008971922099590302)(112, 0.00849157851189375)(113, 0.008660235442221165)(114, 0.008557035587728024)(115, 0.008588016033172607)(116, 0.008735989220440388)(117, 0.008547084406018257)(118, 0.008778677321970463)(119, 0.008687639608979225)(120, 0.00873425230383873)(121, 0.008712735027074814)(122, 0.008525179699063301)(123, 0.008868791162967682)(124, 0.008470195345580578)(125, 0.00856783427298069)(126, 0.008597033098340034)(127, 0.008526703342795372)(128, 0.008490744046866894)(129, 0.008470035158097744)(130, 0.008502052165567875)(131, 0.008502982556819916)(132, 0.008586795069277287)(133, 0.008325262926518917)(134, 0.008646314963698387)(135, 0.008675161749124527)(136, 0.008796455338597298)(137, 0.008442173711955547)(138, 0.008507288992404938)(139, 0.008307352662086487)(140, 0.008281588554382324)(141, 0.008537969551980495)(142, 0.00847679190337658)(143, 0.008241266943514347)(144, 0.008328881114721298)(145, 0.008420298807322979)(146, 0.008387686684727669)(147, 0.008274463936686516)(148, 0.008223549462854862)(149, 0.008449803106486797)(150, 0.008575440384447575)
					
				};
				\addlegendentry{4 0.9-0.95}

			\end{axis}	
		\end{tikzpicture}
		\\Average Loss / Epoch
	\end{minipage}
	\caption{Asymmetric Momentum tested on WRN28-10 with Cifar100. We still only made a swap in the momentum direction, yet it resulted in unexpected effects. The red represents the direction of $\Theta_n$ non-sparse acceleration. We can observe that it experienced significant oscillations in the mid-term, and the final result was also unsatisfactory. Here, the acceleration in the $\Theta_s$ sparse direction showed the best results.}
	\label{Fig0402}
\end{figure*}
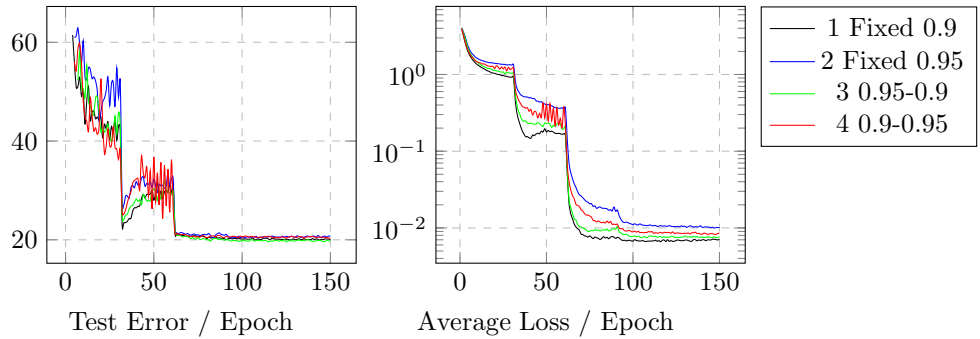

In the first and second baseline group, unlike in Cifar10 where it's quite evident, we can essentially consider both fixed momentum training to be without anomalies lines. However, from the training results, it's easy to see that the outcome with a momentum of 0.95 is not ideal. This is because the weights changes too quickly, causing SGD to be unable to adjust its direction in time.

In the third group, we obtained the best results. We can deduce that in Cifar100, pushing the weights towards the sparse side yields better outcomes, implying that Cifar100 is a sparse dataset, consistent with the structure of the dataset.

However, after \textbf{only swap momentum directions}, we noticed that the fourth group experienced oscillations in the mid-training phase. This result is the most confused. When we intentionally pushed $\Theta_n$ to the left, it was pushed too far, causing a significant oscillation in the overall weights. This is the tug-of-war between the traction produced by the learning rate in the spring model and the pushing force generated by the Asymmetric Momentum. This further verifies the correctness of our theory.

\subsection{Final Training}


To verify if our method can be applied in regular training, we implemented a very simple scheduler. As shown in Algorithm.\ref{alg1}, we maintain a learning rate of 0.1 for the first 30 epochs, and thereafter, for each iteration, we reduce the learning rate to 0.99985 times its previous value. Although the mechanism is empirical, based on many scheduling methods, we believe that reducing the learning rate at every iteration can increase the probability of weights escaping from local minima.

Of course, we also implemented both fixed momentum and asymmetric momentum for comparison. For Cifar10, we chose $\Theta_s$ phase to be 0.9 and $\Theta_n$ phase to be 0.95, aligning with the needs of a non-sparse dataset. For Cifar100, we selected $\Theta_s$ phase to be 0.93 and $\Theta_n$ phase to be 0.9, to cater to the needs of a sparse dataset.

\begin{figure*}[h]
	\centering
	\begin{minipage}[h]{0.4\textwidth}
		\centering
		\begin{tikzpicture} 
			\begin{axis}[
				scale=0.6,
				tick align=inside, 
				xtick = {50,100,150},
				xmin = 45,
				ymin = 3.8,
				ytick = {4,4.5,5,5.5,6},
				grid style={dashed},
				xmajorgrids =true,
				ymajorgrids =true,
				legend style={at={(1,0.8)},anchor=east} 
				]
				
					%
				\addplot[smooth,black] plot coordinates { 
					(68, 5.599999904632568)(69, 5.78000020980835)(70, 5.309999942779541)(71, 5.079999923706055)(72, 4.960000038146973)(73, 5.099999904632568)(74, 5.389999866485596)(75, 5.059999942779541)(76, 5.010000228881836)(77, 5.0)(78, 4.670000076293945)(79, 4.630000114440918)(80, 4.579999923706055)(81, 4.5)(82, 4.570000171661377)(83, 4.760000228881836)(84, 4.590000152587891)(85, 4.510000228881836)(86, 4.460000038146973)(87, 4.489999771118164)(88, 4.409999847412109)(89, 4.409999847412109)(90, 4.539999961853027)(91, 4.449999809265137)(92, 4.380000114440918)(93, 4.380000114440918)(94, 4.389999866485596)(95, 4.440000057220459)(96, 4.389999866485596)(97, 4.320000171661377)(98, 4.28000020980835)(99, 4.369999885559082)(100, 4.320000171661377)(101, 4.260000228881836)(102, 4.260000228881836)(103, 4.190000057220459)(104, 4.170000076293945)(105, 4.190000057220459)(106, 4.260000228881836)(107, 4.340000152587891)(108, 4.119999885559082)(109, 4.170000076293945)(110, 4.110000133514404)(111, 4.179999828338623)(112, 4.139999866485596)(113, 4.170000076293945)(114, 4.130000114440918)(115, 4.150000095367432)(116, 4.150000095367432)(117, 4.21999979019165)(118, 4.179999828338623)(119, 4.130000114440918)(120, 4.110000133514404)(121, 4.230000019073486)(122, 4.179999828338623)(123, 4.179999828338623)(124, 4.199999809265137)(125, 4.150000095367432)(126, 4.210000038146973)(127, 4.190000057220459)(128, 4.300000190734863)(129, 4.300000190734863)(130, 4.239999771118164)(131, 4.21999979019165)(132, 4.190000057220459)(133, 4.340000152587891)(134, 4.210000038146973)(135, 4.239999771118164)(136, 4.130000114440918)(137, 4.260000228881836)(138, 4.199999809265137)(139, 4.239999771118164)(140, 4.21999979019165)(141, 4.170000076293945)(142, 4.25)(143, 4.210000038146973)(144, 4.269999980926514)(145, 4.170000076293945)(146, 4.230000019073486)(147, 4.150000095367432)(148, 4.179999828338623)(149, 4.179999828338623)(150, 4.269999980926514)
					
				};
				
				\addplot[smooth,green] plot coordinates { 
					(73, 5.53000020980835)(74, 5.46999979019165)(75, 5.349999904632568)(76, 4.989999771118164)(77, 5.210000038146973)(78, 4.860000133514404)(79, 4.900000095367432)(80, 4.510000228881836)(81, 4.690000057220459)(82, 4.760000228881836)(83, 4.599999904632568)(84, 4.489999771118164)(85, 4.389999866485596)(86, 4.25)(87, 4.300000190734863)(88, 4.230000019073486)(89, 4.150000095367432)(90, 4.25)(91, 4.079999923706055)(92, 4.03000020980835)(93, 4.050000190734863)(94, 4.119999885559082)(95, 4.03000020980835)(96, 4.059999942779541)(97, 4.239999771118164)(98, 3.990000009536743)(99, 4.070000171661377)(100, 4.090000152587891)(101, 4.019999980926514)(102, 3.990000009536743)(103, 3.9800000190734863)(104, 3.9700000286102295)(105, 4.050000190734863)(106, 4.019999980926514)(107, 4.03000020980835)(108, 3.9200000762939453)(109, 3.930000066757202)(110, 4.019999980926514)(111, 3.990000009536743)(112, 3.950000047683716)(113, 4.050000190734863)(114, 3.9700000286102295)(115, 4.159999847412109)(116, 3.9600000381469727)(117, 4.0)(118, 3.9700000286102295)(119, 4.010000228881836)(120, 3.990000009536743)(121, 3.990000009536743)(122, 4.099999904632568)(123, 4.010000228881836)(124, 3.9600000381469727)(125, 3.9600000381469727)(126, 3.940000057220459)(127, 3.950000047683716)(128, 4.039999961853027)(129, 4.03000020980835)(130, 4.019999980926514)(131, 4.039999961853027)(132, 4.0)(133, 4.0)(134, 4.019999980926514)(135, 3.9200000762939453)(136, 4.0)(137, 4.019999980926514)(138, 3.9800000190734863)(139, 3.9700000286102295)(140, 3.9800000190734863)(141, 3.9600000381469727)(142, 3.940000057220459)(143, 3.9800000190734863)(144, 4.0)(145, 4.0)(146, 3.9200000762939453)(147, 4.0)(148, 3.9800000190734863)(149, 3.9800000190734863)(150, 4.010000228881836)
					
				};
				
			\end{axis}	
		\end{tikzpicture}
		\\Cifar10
	\end{minipage}\begin{minipage}[h]{0.4\textwidth}
		\centering
		\begin{tikzpicture} 
			\begin{axis}[
				scale=0.6,
				tick align=inside, 
				xtick = {50,100,150},
				xmin = 45,
				ymin = 18.5,
				ytick = {19,20,21,22},
				grid style={dashed},
				xmajorgrids =true,
				ymajorgrids =true,
				legend style={at={(1.5,1)},anchor=north} 
				]
				
					%
				\addplot[smooth,black] plot coordinates { 
					(65, 21.549999237060547)(66, 21.809999465942383)(67, 20.979999542236328)(68, 21.969999313354492)(69, 20.579999923706055)(70, 20.34000015258789)(71, 20.219999313354492)(72, 19.809999465942383)(73, 19.93000030517578)(74, 19.899999618530273)(75, 19.979999542236328)(76, 19.739999771118164)(77, 19.530000686645508)(78, 19.56999969482422)(79, 19.290000915527344)(80, 19.780000686645508)(81, 19.1200008392334)(82, 19.360000610351562)(83, 19.56999969482422)(84, 19.360000610351562)(85, 19.280000686645508)(86, 19.25)(87, 19.309999465942383)(88, 19.489999771118164)(89, 19.489999771118164)(90, 19.6200008392334)(91, 19.329999923706055)(92, 19.68000030517578)(93, 19.6299991607666)(94, 19.309999465942383)(95, 19.399999618530273)(96, 19.420000076293945)(97, 19.420000076293945)(98, 19.280000686645508)(99, 19.3700008392334)(100, 19.440000534057617)(101, 19.56999969482422)(102, 19.31999969482422)(103, 19.25)(104, 19.31999969482422)(105, 19.209999084472656)(106, 19.299999237060547)(107, 19.34000015258789)(108, 19.399999618530273)(109, 19.40999984741211)(110, 19.440000534057617)(111, 19.329999923706055)(112, 19.31999969482422)(113, 19.420000076293945)(114, 19.15999984741211)(115, 19.200000762939453)(116, 19.290000915527344)(117, 19.3700008392334)(118, 19.25)(119, 19.200000762939453)(120, 19.34000015258789)(121, 19.3799991607666)(122, 19.139999389648438)(123, 19.209999084472656)(124, 19.209999084472656)(125, 19.1299991607666)(126, 19.200000762939453)(127, 19.31999969482422)(128, 19.3799991607666)(129, 19.420000076293945)(130, 19.3700008392334)(131, 19.229999542236328)(132, 19.489999771118164)(133, 19.350000381469727)(134, 19.350000381469727)(135, 19.31999969482422)(136, 19.399999618530273)(137, 19.18000030517578)(138, 19.209999084472656)(139, 19.399999618530273)(140, 19.270000457763672)(141, 19.360000610351562)(142, 19.450000762939453)(143, 19.25)(144, 19.31999969482422)(145, 19.34000015258789)(146, 19.200000762939453)(147, 19.190000534057617)(148, 19.3799991607666)(149, 19.489999771118164)(150, 19.290000915527344)
					
				};
				\addlegendentry{1 Fixed 0.9}
				
				\addplot[smooth,green] plot coordinates { 
					(68, 21.559999465942383)(69, 21.6299991607666)(70, 21.270000457763672)(71, 21.670000076293945)(72, 20.979999542236328)(73, 20.399999618530273)(74, 20.170000076293945)(75, 19.90999984741211)(76, 19.719999313354492)(77, 19.6299991607666)(78, 19.280000686645508)(79, 19.43000030517578)(80, 19.559999465942383)(81, 19.309999465942383)(82, 19.350000381469727)(83, 18.969999313354492)(84, 19.170000076293945)(85, 19.079999923706055)(86, 19.170000076293945)(87, 18.860000610351562)(88, 19.459999084472656)(89, 19.010000228881836)(90, 19.15999984741211)(91, 19.110000610351562)(92, 19.170000076293945)(93, 19.219999313354492)(94, 19.020000457763672)(95, 18.81999969482422)(96, 19.209999084472656)(97, 19.049999237060547)(98, 19.030000686645508)(99, 18.799999237060547)(100, 18.899999618530273)(101, 19.09000015258789)(102, 18.989999771118164)(103, 19.0)(104, 19.06999969482422)(105, 18.860000610351562)(106, 18.920000076293945)(107, 18.829999923706055)(108, 18.81999969482422)(109, 18.829999923706055)(110, 18.790000915527344)(111, 18.889999389648438)(112, 18.93000030517578)(113, 18.889999389648438)(114, 18.959999084472656)(115, 18.90999984741211)(116, 19.06999969482422)(117, 18.860000610351562)(118, 18.959999084472656)(119, 18.950000762939453)(120, 18.979999542236328)(121, 19.09000015258789)(122, 18.8700008392334)(123, 19.030000686645508)(124, 18.989999771118164)(125, 18.989999771118164)(126, 19.0)(127, 19.170000076293945)(128, 18.950000762939453)(129, 19.149999618530273)(130, 19.139999389648438)(131, 18.8799991607666)(132, 18.90999984741211)(133, 19.1200008392334)(134, 19.020000457763672)(135, 19.030000686645508)(136, 19.020000457763672)(137, 19.149999618530273)(138, 18.920000076293945)(139, 19.049999237060547)(140, 19.1200008392334)(141, 18.969999313354492)(142, 19.25)(143, 19.110000610351562)(144, 19.110000610351562)(145, 18.93000030517578)(146, 19.1200008392334)(147, 19.299999237060547)(148, 18.93000030517578)(149, 19.040000915527344)(150, 19.049999237060547)
					
				};
				\addlegendentry{2 LCAM}
				
			\end{axis}	
		\end{tikzpicture}
		\text{Cifar100}
	\end{minipage}
	\caption{Final results on WRN28-10 with Cifar10 and Cifar100. The vertical axis represents the test error rate, while the horizontal axis represents the epoch.}
	\label{Fig0403}
\end{figure*}
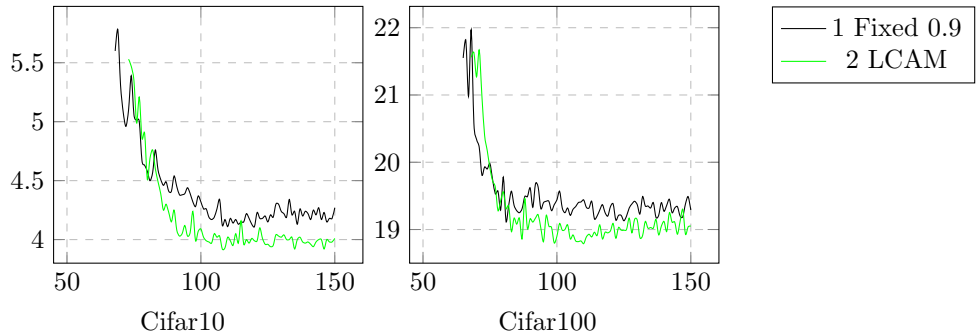

It should be noted that the green line does not represent a consistent outcome. This is due to the influence of local minima, causing its final convergence location to be unstable. The final results, as depicted in the Figures.\ref{Fig0403} and Tables.\ref{tabel1}, demonstrate that by the epoch at 120, we can achieve accuracy comparable to or better than that in other papers by the epoch at 200(a lower test error rate).

\begin{table}[h]
	\begin{center}
		\caption{Final results compared with WRN original paper and Cosine Annealing Scheduler\cite{loshchilov2016sgdr}. We can significantly reduce the required epochs while achieving a slightly better accuracy.}
		\label{tabel1}
		\begin{tabular}{c|c|c|c}
			\hline
			Test Error (\%)&Epochs& Cifar10 & Cifar100\\
			\hline
			WRN&200& 4.00 & 19.25\\
			Cosine Annealing&200& 4.03 & 19.58\\
			\textbf{Ours LCAM}&\textbf{120} & \textbf{3.98} & \textbf{19.22 (18.92 in 50\%)}\\
			\hline
		\end{tabular}
	\end{center}
\end{table}

\section{Conclusion}
Although the experiment is simple, the phenomena are obvious. We explain why adaptive optimizers only suit for sparse gradient, prove that within the model weights, the gradient has specificity on various directions based on the dataset sparsity. We introduced the theories of Weight Coupling and Weight Traction and utilized this mechanism through Asymmetric Momentum. By comparing the current loss value with the average loss value after each training iteration, we segmented the training process into two phases with different momentum towards sparse or non-sparse direction, and can be adapted for different datasets. In training, Loss-Controlled Asymmetric Momentum(LCAM) retains the benefits of traditional SGD, achieving the best accuracy with minimal training resources required for each iteration. Additionally, the demand epochs for training can be nearly halved.

\section{Reproducibility Statement}
Figure.\ref{Fig0401} and Figure.\ref{Fig0402} are our main experiments, validating the correctness of our theory. The abnormal phenomena in the specially designed experiment is merely affected by the local minima, leading to extremely high experimental stability and 100\% Reproducibility.

However, it needs to be stated that, in Figure.\ref{Fig0403}, the final test error is significantly affected by the local minima, which is not the part we focus on. After extensive experimentation, compared with a fixed momentum of 0.9, the probability of improvement by our method is roughly 80\% in Cifar10, 50\% in Cifar100. This implies that choosing a more appropriate scheduler to replace our simple one could potentially yield better stability.


\begin{thebibliography}{1}
	
\bibitem{duchi2011adaptive}
Duchi, John and Hazan, Elad and Singer, Yoram.
\newblock Adaptive subgradient methods for online learning and stochastic optimization.
\newblock Journal of machine learning research, 2011, 12(7).

\bibitem{RmsProp}
T. Tieleman and G. Hinton.
\newblock RmsProp: Divide the gradient by a running average of its recent mag- nitude.
\newblock COURSERA: Neural Networks for Machine Learning, 2012.

\bibitem{kingma2014adam}
Kingma, Diederik P and Ba, Jimmy.
\newblock Adam: A method for stochastic optimization.
\newblock arXiv preprint arXiv:1412.6980, 2014.

\bibitem{reddi2018convergence}
Reddi, Sashank J and Kale, Satyen and Kumar, Sanjiv.
\newblock On the Convergence of Adam and Beyond.
\newblock International Conference on Learning Representations, 2018.

\bibitem{loshchilov2018decoupled}
Loshchilov, Ilya and Hutter, Frank.
\newblock Decoupled Weight Decay Regularization.
\newblock International Conference on Learning Representations, 2018.

\bibitem{dauphin2014identifying}
Dauphin Y N, Pascanu R, Gulcehre C, et al.
\newblock Identifying and attacking the saddle point problem in high-dimensional non-convex optimization.
\newblock Advances in neural information processing systems, 2014, 27.

\bibitem{robbins1951stochastic}
Robbins, Herbert and Monro, Sutton.
\newblock A stochastic approximation method.
\newblock The annals of mathematical statistics.

\bibitem{loshchilov2016sgdr}
Loshchilov, Ilya and Hutter, Frank.
\newblock Sgdr: Stochastic gradient descent with warm restarts.
\newblock International Conference on Learning Representations. 2016.

\bibitem{zagoruyko2016wide}
Zagoruyko, Sergey and Komodakis, Nikos.
\newblock Wide residual networks.
\newblock Procedings of the British Machine Vision Conference 2016.

\bibitem{he2016deep}
He, Kaiming and Zhang, Xiangyu and Ren, Shaoqing and Sun, Jian.
\newblock Deep residual learning for image recognition.
\newblock Proceedings of the IEEE conference on computer vision and pattern recognition. 770--778, 2016.

\bibitem{krizhevsky2009learning}
Krizhevsky, Alex and Hinton, Geoffrey and others.
\newblock Learning multiple layers of features from tiny images.
\newblock Citeseer, 2009.

\bibitem{keskar2017improving}
Keskar, Nitish Shirish and Socher, Richard.
\newblock Improving generalization performance by switching from adam to sgd.
\newblock arXiv preprint arXiv:1712.07628, 2017.
	
	
\end{thebibliography}

%

\end{document}